\documentclass[10pt,twocolumn,letterpaper]{article}
\usepackage{cvpr}              % To produce the CAMERA-READY version
\usepackage{multirow}
\usepackage{textcomp}
\usepackage[mode=buildnew]{standalone}
\usepackage{lineno}
\usepackage{tikz}
\usepackage{colortbl}
\usepackage{pgfplots}
\pgfplotsset{compat=1.18}
\definecolor{cvprblue}{rgb}{0.21,0.49,0.74}
\usepackage[pagebackref,breaklinks,colorlinks,allcolors=cvprblue]{hyperref}
\usepackage[accsupp]{axessibility}
\def\paperID{14378} % *** Enter the Paper ID here
\def\confName{CVPR}
\def\confYear{2025}

\begin{document}
%%%%%%%%% TITLE - PLEASE UPDATE
\title{Multi-View Pose-Agnostic Change Localization with Zero Labels}

%%%%%%%%% AUTHORS - PLEASE UPDATE
\author{Chamuditha Jayanga Galappaththige$^{1,2}$ \hspace{10pt} Jason Lai$^{3}$ \hspace{10pt}  Lloyd Windrim$^{2,4}$ \\ Donald Dansereau$^{2,3}$  \hspace{10pt}   Niko S\"underhauf$^{1,2}$ \hspace{10pt}  Dimity Miller$^{1,2}$ \\
$^1$Queensland University of Technology \hspace{1pt} $^2$ARIAM$^*$ \hspace{1pt} $^3$ACFR, University of Sydney \hspace{1pt} $^4${Abyss Solutions}
% Institution1 address 
\\
{\tt\small \{chamuditha.galappaththige, d24.miller\}@.qut.edu.au}
% For a paper whose authors are all at the same institution,
% omit the following lines up until the closing ``}''.
% Additional authors and addresses can be added with ``\and'',
% just like the second author.
% To save space, use either the email address or home page, not both
% \and
% Second Author\\
% Institution2\\
% First line of institution2 address\\
% {\tt\small secondauthor@i2.org}
}
\maketitle
\def \thefootnote{*}\footnotetext{This work was supported by the ARC Research Hub in Intelligent Robotic Systems for Real-Time Asset Management (ARIAM) (IH210100030) and Abyss Solutions. C.J., N.S., and D.M. also acknowledge ongoing support from the QUT Centre for Robotics.}
\begin{abstract}
Autonomous agents often require accurate methods for detecting and localizing changes in their environment, particularly when observations are captured from unconstrained and inconsistent viewpoints. We propose a novel label-free, pose-agnostic change detection method that integrates information from multiple viewpoints to construct a change-aware 3D Gaussian Splatting (3DGS) representation of the scene. With as few as 5 images of the post-change scene, our approach can learn an additional change channel in a 3DGS and produce change masks that outperform single-view techniques. Our change-aware 3D scene representation additionally enables the generation of accurate change masks for unseen viewpoints. Experimental results demonstrate state-of-the-art performance in complex multi-object scenes, achieving a \textbf{1.7$\times$} and \textbf{1.5$\times$} improvement in Mean Intersection Over Union and F1 score respectively over other baselines. 
We also contribute a new real-world dataset to benchmark change detection in diverse challenging scenes in the presence of lighting variations.
Our code and the dataset are available at \href{https://chumsy0725.github.io/MV-3DCD/}{MV-3DCD.github.io}.

\end{abstract}    
\section{Introduction}
\label{sec:intro}

\begin{figure}[t]
    \centering
   \includegraphics[width=\linewidth,trim={8.4cm 5cm 8cm 5cm}]{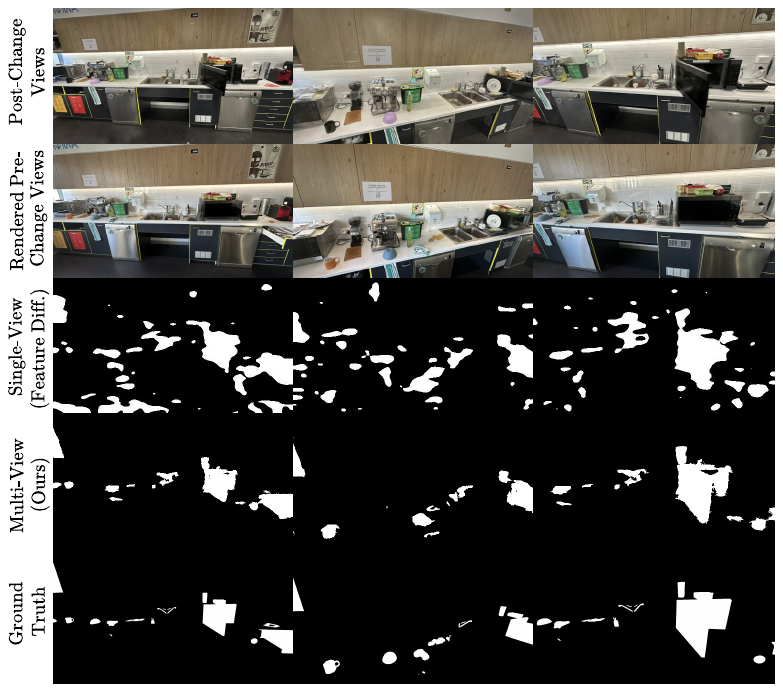}
   \caption{Our multi-view approach to visual change detection (second row from bottom) enforces consistency of the predicted changes across multiple viewpoints by embedding change information in a 3D Gaussian Splatting model of the scene. This effectively suppresses many of the false-positive detections exhibited by current single-view methods (middle row).}
    \label{fig:hero}
\end{figure}

There is increasing effort to develop autonomous agents that assist us with complex tasks, from handling daily chores to performing undesirable work. Capable autonomous agents require the ability to detect and interpret changes in their environment, enabling them to update maps and re-plan tasks or perform applied tasks such as infrastructure or environment monitoring. Change detection remains a challenging task in 3D scenes, particularly when an agent observes the scene from two sets of views that have no constraint on the poses (i.e. consider a robot that captures images of a scene following a random trajectory at each inspection round). 
% different angles at different times.

Many established change detection methods rely on precise alignment between a pre-change image and post-change image to localize the change ~\cite{jhamtani_learning_2018,chen_spatial-temporal_2020,alcantarilla_street-view_2018,caye_daudt_fully_2018}, limiting their applicability to scenes without viewpoint consistency. Some approaches extend to detect changes in images with inconsistent viewpoints~\cite{lin_robust_2024,lee_semi-supervised_2024,sachdeva_change_2023}, but learn viewpoint invariance by training on image pairs labeled with changes and showcasing viewpoint discrepancy. Supervised learning can have limitations for change detection, including the cost of labeling datasets and significant performance drops under distribution shift (such as environments not present in the dataset)~\cite{jhamtani_learning_2018,alcantarilla_street-view_2018,varghese_changenet_2019,wang2022c3po}. In this paper, we address the problem of label-free, pose-agnostic change localization, where changes are detected between a pre-change scene and a post-change scene, without labeled data for training or aligned viewpoints for observations between the scenes.

Recent works~\cite{zhou2024pad,kruse2024splatpose} perform label-free, pose-agnostic change localization by learning a 3D representation of the scene, such as a Neural Radiance Field (NeRF)~\cite{mildenhall_nerf_2021} or 3D Gaussian Splatting (3DGS)~\cite{kerbl3Dgaussians}, and rendering images from the viewpoints of observed images. Changes are detected through feature-level comparisons between the observed and rendered images when using a pre-trained vision model~\cite{kruse2024splatpose,zhou2024pad}. While this is a feasible approach to pose-agnostic change detection, such approaches struggle to produce accurate change maps in the presence of view-dependent feature-level inconsistencies (e.g. reflections, shadows, unseen regions) common in real-world scenarios.

For the first time, we propose a novel \emph{multi-view} change detection method that is both pose-agnostic and label-free. Our approach integrates change information from multiple viewpoints by constructing a 3DGS model of the environment, encoding not only appearance but also a measure of \emph{change} (an explicit 3D representation of change). This enables the generation of change masks for any viewpoint in the scene, including those not yet observed post-change. By leveraging multiple viewpoints and incorporating change masks that are both feature- and structure-aware, our method produces robust multi-view change masks, mitigating potential view-dependent false changes flagged at the feature level (see Fig.~\ref{fig:hero}). Furthermore, we show that our change-aware 3DGS can serve as a multi-view extension for \emph{any} change mask generation method (see Sec.~\ref{sec:scd}).

We make three key claims that are supported by our experiments: First, our approach achieves state-of-the-art performance, particularly in complex multi-object scenes. Second, our change-aware 3D scene representation allows us to generate change predictions for entirely unseen views in the post-change scene, which current methods are unable to do. Third, pre-trained features and the Structural Similarity Index Measure (SSIM)~\cite{wang2004image} contain complemental change information, and their combination generates robust change masks to learn a change-aware 3DGS.

We additionally contribute a novel dataset encompassing 10 real-world scenes with multiple objects and diverse changes. Our dataset includes variations in lighting, indoor and outdoor settings, and multi-perspective captures, enabling a finer-grained analysis of change detection methods in realistic conditions. We evaluate our approach on three change detection datasets including our novel dataset, comparing to existing state-of-the-art methods and demonstrating significant improvements in performance.

\section{Related Work}
\subsection{Pair-wise (2D) Scene Change Detection}

A typical change detection scenario involves a pair of before-and-after RGB images without explicitly considering a 3D scene~\cite{alcantarilla_street-view_2018, sachdeva_change_2023-1, bandara_transformer-based_2022, sakurada_change_2015, li_cognitive_2020, park_changesim_2021}. These images often adhere to specific conditions: the camera remains fixed, resulting in images related by an identity transform, as in surveillance footage~\cite{jhamtani_learning_2018, krajnik_long-term_2014}; the scene is planar, as in bird's-eye view or satellite images~\cite{chen_spatial-temporal_2020, caye_daudt_fully_2018}; or there is minimal viewpoint shift, as in street-view scenes capturing distant buildings or objects~\cite{alcantarilla_street-view_2018, sakurada_change_2015}. In these cases, models are generally expected to learn to identify changes between image pairs by localizing differences through segmentation~\cite{sakurada_weakly_2020, bergmann_mvtec_2019, alcantarilla_street-view_2018}.

Convolutional Neural Networks (CNNs) have been widely studied for localizing changes~\cite{Long_2015_CVPR, khan_learning_2017, caye_daudt_fully_2018, varghese_changenet_2019, sakurada_weakly_2020, wang2022c3po}. More recently, transformer-based architectures~\cite{dosovitskiy2021an} have shown the ability to learn rich, context-aware representations through attention mechanisms, advancing change detection tasks~\cite{vaswani_attention_2017, Wang:21, bandara_transformer-based_2022, shi_divided_2022, fang_changer_2023}. Foundation models, such as DINOv2~\cite{oquab2023dinov2}, have proven to be robust pre-trained backbones for feature extraction, enhancing change detection across diverse applications~\cite{martinson_meaningful_2024, lin_robust_2024}.

\subsection{2D-3D Scene-level Change Detection}
2D to 3D scene-level change detection tackles the challenging and realistic task of identifying changes in 3D scenes, where large viewpoint shifts, severe occlusions, and disocclusions are common. While detecting changes in 3D scenes from sparse 2D RGB images remains underexplored, Sachdeva \etal~\cite{sachdeva_change_2023} recently introduced a ``register-and-difference'' approach that leverages frozen embeddings from a pre-trained backbone and feature differences to detect changes. Similarly, Lin \etal~\cite{lin_robust_2024} proposed a cross-attention mechanism built on DINOv2~\cite{oquab2023dinov2} to address viewpoint inconsistencies in street-view settings. However, both methods rely solely on image-to-image comparisons and do not explicitly construct a 3D representation of the scene.

Related to scene-level change detection is pose-agnostic anomaly detection. Anomaly detection typically leverages unsupervised learning to build a normality model from a set of 2D images, tagging images inconsistent with this model as anomalies during inference \cite{zhang_realnet_2024,zavrtanik_draem_2021,liang_omni-frequency_2023,zavrtanik_dsr_2022}. Recently, Zhou \etal~\cite{zhou2024pad} introduced a pose-agnostic anomaly detection dataset consisting of small-scale scenes containing single toy LEGO objects. Closely related to our work, OmniPoseAD~\cite{zhou2024pad} and SplatPose~\cite{kruse2024splatpose} explore this dataset to build 3D object representations of a scene containing a faultless object. OmniPoseAD employs NeRFs~\cite{mildenhall2020nerf} to model the object, using coarse-to-fine pose estimation with iNeRF~\cite{yen2020inerf} to render a matching viewpoint and generates anomaly scores by comparing multi-scale features from a pre-trained CNN. SplatPose replaces NeRF with 3DGS~\cite{kerbl3Dgaussians} and directly learns rigid transformations for each Gaussian, bypassing iNeRF. Both methods leverage a 3D scene representation, but only consider anomaly detection on a single per-view image basis -- we extend beyond these works by leveraging multiple views and the 3D scene representation to learn more robust multi-view change masks.

\subsection{Learning a 3D Representation} \label{sec:related_representations}

Learning a 3D representation of a scene has been used by prior works to enable pose-agnostic, unsupervised change detection~\cite{zhou2024pad, kruse2024splatpose}. 
Complex geometries can be represented as continuous implicit fields using coordinate-based neural networks. For example, signed distance fields~\cite{park_deepsdf_2019,takikawa_neural_2021} capture the distance of each point to object surfaces, while occupancy networks~\cite{mescheder_occupancy_2019} indicate whether points lie within an object. Recent advances in high-fidelity scene representations, such as NeRFs~\cite{mildenhall_nerf_2021} and variants~\cite{muller_instant_2022, barron_mip-nerf_2022, fridovich-keil_plenoxels_2022}, model scenes by regressing a 5D plenoptic function~\cite{landy_plenoptic_1991}, outputting view-independent density and view-dependent radiance for photorealistic novel view synthesis.

In contrast to implicit fields, 3DGS~\cite{kerbl3Dgaussians} provides an explicit scene representation using anisotropic 3D Gaussians, enabling high-quality, real-time novel view synthesis. Each Gaussian is defined by a center position \( \mu \) and covariance matrix \( \Sigma \), calculated from a scaling matrix \( S \) and rotation matrix \( R \) as \( \Sigma = R S S^T R^T \). Additionally, an opacity factor \( \alpha \) and color component \( c \), modeled with spherical harmonics, are learned to capture view-dependent appearance. To initialize, 3DGS uses Structure-from-Motion (SfM) with COLMAP~\cite{schoenberger2016sfm} to estimate camera poses and create a sparse point cloud from multi-view images. Gaussian parameters and color components are then optimized by comparing rendered views with ground truth images using a combination of \( L_1 \) loss and a D-SSIM loss term~\cite{kerbl3Dgaussians}.

\section{Methodology}
\begin{figure*}[t!]
    \centering
    \includegraphics[width=\textwidth]{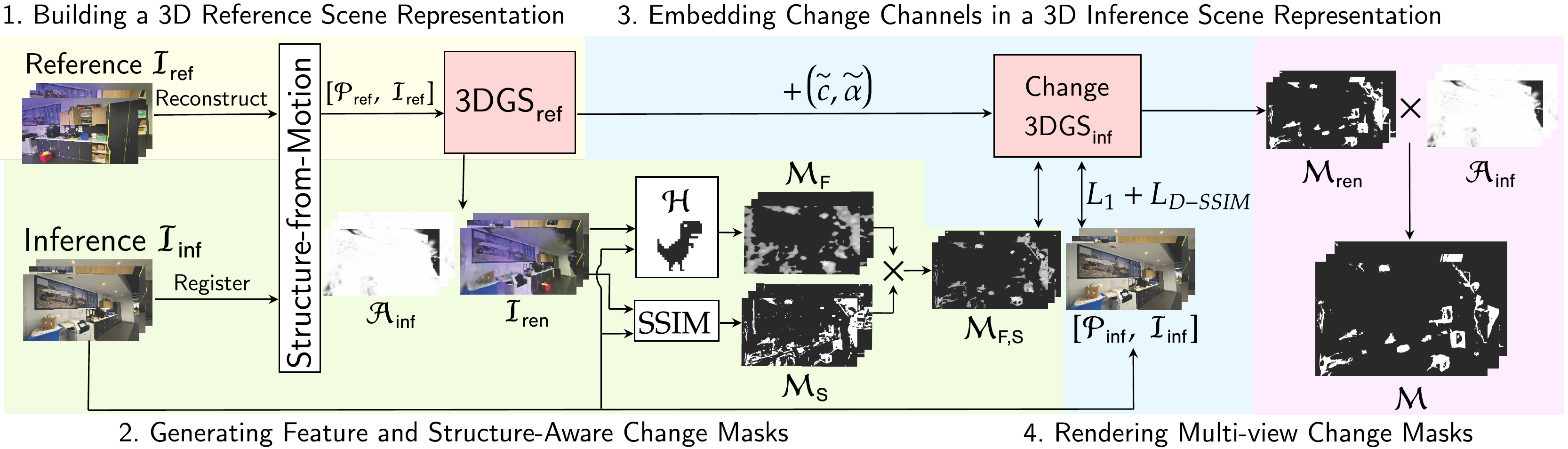}
    \caption{An overview of our proposed approach for multi-view pose-agnostic change detection. We leverage a 3DGS representation of the pre-change (\emph{reference}) scene to build feature and structure-aware change masks given images of the post-change (\emph{inference)} scene. We embed this information as additional change channels into the representation, which can be used to render multi-view change masks.}
    \label{fig:overview}
\end{figure*}

An overview of our proposed \emph{multi-view} change detection approach is shown in Fig.~\ref{fig:overview}. We construct a 3DGS~\cite{kerbl3Dgaussians} representation for the pre-change (\emph{reference}) scene, allowing us to render pre-change images from novel viewpoints (Sec.~\ref{sec:method_build_scene}). After collecting images from the post-change (\emph{inference}) scene, we compare to corresponding rendered pre-change images and compute feature and structure-aware change masks (Sec.~\ref{sec:method_changemasks}). We then learn an \emph{updated} 3DGS for the post-change scene that also embeds Gaussian-specific change channels for reconstructing change masks, leveraging the multiple views from the 3D scene (Sec.~\ref{sec:method_changechannels}). This \emph{change-aware 3DGS} can be queried for any pose to generate a multi-view change mask of the scene (Sec.~\ref{sec:method_rendermasks}). We additionally introduce a data augmentation strategy to increase the number of change masks used to learn our change-aware 3DGS (Sec.~\ref{sec:method_dataaugment}).

\subsection{Problem Setup}
A set of $n_{\text{ref}}$ images are collected from a reference scene, \( \mathcal{I}_{\text{ref}} = \{I_{\text{ref}}^k\}_{k=1}^{n_{\text{ref}}} \). Changes in this scene then occur, including structural changes (addition, removal, or movement of objects) and surface-level changes (changes to texture or color of objects, drawings on surfaces). ``Distractor'' or irrelevant visual changes can also occur, such as changes in lighting, shadows, or reflections in the scene. A set of $n_{\text{inf}}$ images are collected from the scene post-change, referred to as the inference scene, \( \mathcal{I}_{\text{inf}} = \{I_{\text{inf}}^k\}_{k=1}^{n_{\text{inf}}} \). Our objective is to generate a set of segmentation masks \( \mathcal{M} = \{M^{k}\}_{k=1}^{n_{\text{inf}}} \) for all images in \( \mathcal{I}_{\text{inf}} \) that localizes all relevant changes between the reference and inference scenes while disregarding distractor changes.

\subsection{Building a 3D Reference Scene Representation}
\label{sec:method_build_scene}
 
Given the reference scene images \( \mathcal{I}_{\text{ref}} \), we utilise COLMAP~\cite{schoenberger2016sfm} to perform SfM and obtain camera poses for all images, \( \mathcal{P}_{\text{ref}} = \{P_{\text{ref}}^k\}_{k=1}^{n_{\text{ref}}} \). We then use $\mathcal{P}_{\text{ref}}$ and \(  \mathcal{I}_{\text{ref}} \) to construct a 3DGS representation of the reference scene, $\text{3DGS}_{\text{ref}}$, following the pipeline described in ~\cite{kerbl3Dgaussians}. We assume that the number, quality and viewpoints of images in $\mathcal{I}_{\text{ref}}$ is sufficient to build a 3DGS \cite{kerbl3Dgaussians} representation.

\subsection{Generating Feature and Structure-Aware Change Masks}
\label{sec:method_changemasks}
Given the inference scene images \( \mathcal{I}_{\text{inf}} \), we acquire corresponding camera poses \( \mathcal{P}_{\text{inf}} = \{P_{\text{inf}}^k\}_{k=1}^{n_{\text{inf}}} \) by registering \( \mathcal{I}_{\text{inf}} \) to the same SfM reconstruction built from \( \mathcal{I}_{\text{ref}} \) using COLMAP~\cite{schoenberger2016sfm}. This ensures \( \mathcal{P}_{\text{ref}} \) and \( \mathcal{P}_{\text{inf}} \) share a reference frame, assuming that the magnitude of appearance change is not so severe that COLMAP~\cite{schoenberger2016sfm} is unable to make the registration (i.e. inference scene is extremely dark).

We then render a new image set, \( \mathcal{I}_{\text{ren}}\), from our $\text{3DGS}_{\text{ref}}$ with the exact poses from our inference scene \( \mathcal{P}_{\text{inf}} \). Comparing images from \( \mathcal{I}_{\text{ren}}\) with the corresponding pose-aligned image in \( \mathcal{P}_{\text{inf}} \), we can now generate change masks.

\noindent\textbf{Feature-Aware Change Mask: }
We extract a feature-aware change mask by leveraging a pre-trained visual foundation model \( \mathcal{H} \) (specifically DINOv2 \cite{oquab2023dinov2}). We test \( \mathcal{H} \) with \( \mathcal{I}_{\text{ren}}\) 
 and \( \mathcal{I}_{\text{inf}}\) to produce a dense feature set \( \{(f_{\text{ren}}^k, f_{\text{inf}}^k)\}_{k=1}^{n_{\text{inf}}} \) for each pose-aligned image pair. These feature maps are defined by the image height \( h \), width \( w \), patch size \( s \) of the foundation model, and embedding dimension \( d \), \( f \in \mathbb{R}^{\frac{h}{s} \times \frac{w}{s} \times d} \). We then compute a preliminary feature-aware change mask \( D^k \) between \( f_{\text{ren}}^k \) and \( f_{\text{inf}}^k \) across the embedding dimension \( d \) as follows:
\begin{equation}
    \label{eq:feature_diff}
    D^k = \sum_{j=1}^{d} |f_{\text{ren}}^{k,j} - f_{\text{inf}}^{k,j}| \in \mathbb{R}^{\frac{h}{s} \times \frac{w}{s}}.
\end{equation}
We then normalize \( D^k \) values to range between 0 and 1 and apply bicubic interpolation to create a feature-aware change mask with the original image dimensions. We create our final feature-aware change mask, \(M^K_{\text{F}}\), by masking all change values below $0.5$ to equal zero -- this can remove potential low-value false changes flagged in the feature-aware change mask.

\noindent\textbf{Structure-Aware Change Mask: }
Alongside our feature-aware change mask, we additionally generate a structure-aware change mask by leveraging the Structural Similarity Index Measure (SSIM)~\cite{wang2004image}. The SSIM quantifies the similarity between two spatially aligned image signals based on the luminance, contrast, and structure components of the images. It is typically used as a metric for the visual quality of images, for example used in image reconstruction to measure the quality of the reconstruction~\cite{kerbl3Dgaussians}. We observe that the SSIM can also serve as a meaningful measure of change between two images, which is complementary to the feature-level change extracted from a pre-trained model. We generate our structure-aware change masks by applying the SSIM to the pairs of \( \mathcal{I}_{\text{ren}}\) 
 and \( \mathcal{I}_{\text{inf}}\), and binarizing the output to filter for low-similarity, high visual change values,
\begin{equation}
    M^{k}_{\text{S}} = \mathbf{1}(\text{SSIM}(I^k_{\text{ren}}, I^k_{\text{inf}}) \leq 0.5),
\end{equation}
where \( \mathbf{1} \) is the indicator function.

\noindent\textbf{Combined Candidate Change Mask: }
We combine the feature-aware and structure-aware change masks by element-wise multiplication to create the final candidate change masks that filter for detected changes at both the features and pixel level:
\begin{equation}
    M^k_{\text{F,S}} = \{M^k_{\text{F}} \cdot M^k_{\text{S}} \}_{k=1}^{n_{\text{inf}}}. 
\end{equation}

Next, we describe how the individual per-view change masks $M^k_{\text{F, S}}$ are combined and fused through the change-aware 3DGS model -- making our approach \textit{multi-view}.

\subsection{Embedding Change Channels in a 3D Inference Scene Representation}
\label{sec:method_changechannels}
A core contribution of our method is that we move beyond change masks generated by individual images to create change masks that leverage our 3D reference scene representation, i.e. multi-view change masks. We achieve this by learning a new 3DGS representation for the inference scene that also contains change information from our feature and structure-aware change masks $\mathcal{M}_\text{F,S}$. We embed this change information directly into a 3DGS by learning two additional channels per Gaussian -- a change magnitude \( \tilde{c} \) (i.e. the level of change each Gaussian captures in the scene) and a change opacity factor \( \tilde{\alpha} \) (which allows us to model which Gaussians contribute to the pixel change values in \(\mathcal{M}_{\text{F,S}} \), see Supp. Material for further discussion). Using these new change parameters, we can then render a change mask from the 3DGS alongside RGB images using the standard rasterization process~\cite{kerbl3Dgaussians}.

To achieve this, we create a new change-aware 3DGS for the inference scene, ${\text{Change-3DGS}_\text{inf}}$, that is initialized with the learned Gaussians from  ${\text{3DGS}_\text{ref}}$. For each Gaussian, we add an additional two parameters to model change in the scene (\( \tilde{c} \), \( \tilde{\alpha} \)). We then re-optimize ${\text{Change-3DGS}_\text{inf}}$ given \( \mathcal{I}_{\text{inf}} \), \( \mathcal{P}_{\text{inf}} \) and \(\mathcal{M}_{\text{F,S}} \), following the standard optimization pipeline described in ~\cite{kerbl3Dgaussians} while including an additional $L_1$ and D-SSIM loss terms to learn the change channel values. For the best performance of our method, ${\text{Change-3DGS}_\text{inf}}$ is initialized with the pre-trained ${\text{3DGS}_\text{ref}}$ so that Gaussians relating to structural changes in the inference scene are retained (see Supp. Material for an in-depth discussion).

Critically, we model \( \tilde{c} \) using a spherical harmonics coefficient degree of zero. Typically in 3DGS~\cite{kerbl3Dgaussians}, a higher degree (degree 3) of spherical harmonics coefficients is used to model view-dependent color, effectively capturing color variations across different viewing directions. We hypothesize that changes in a scene are largely view-independent and that most view-dependent variations in our change masks arise from false positive change predictions, such as reflections, shadows, or minor misalignment between the rendered and inference images. Under this hypothesis, it is then preferable to model change with a low degree of spherical harmonics coefficients so that we can effectively leverage individual change masks to collectively learn true regions of change in the scene while not overfitting to view-dependent false positive changes -- we confirm this in Sec.~\ref{sec:ablation_sh}.

\subsection{Rendering Multi-View Change Masks}
\label{sec:method_rendermasks}
Given $\text{Change-3DGS}_{\text{inf}}$ and any query pose $P_{\text{query}}$, we can render a multi-view change mask. Given our problem setup, we render change masks for all poses from the inference scene, \( \mathcal{M}_{\text{ren}} = \{M^k_{\text{ren}}\}_{k=1}^{n_{\text{inf}}} \). Notably, our approach allows us to also render change masks for viewpoints that are novel to both the reference and inference scene (see Sec.~\ref{sec:n_views} for a further discussion).

As the reference and inference scenes are collected with random trajectories, independently, it is possible that inference images capture scene regions that were absent in the reference image set. Previously unseen regions of the 3DGS do not contain Gaussians, and thus rendered images of such regions are represented with black pixels (the 3DGS background color). To avoid falsely calculating these unseen areas as changes, we exclude them from the rendered change mask as a final post-processing step.

We render the alpha channel \( \mathcal{A_{\text{ren}}} = \{A_{\text{ren}}^k\}_{k=1}^{n_{\text{inf}}} \) alongside \( \mathcal{I}_{\text{ren}} \) as it provides per-pixel opacity between the foreground pixel versus the background. For unseen regions, a 3DGS renders the unseen region as the background color, resulting in alpha channel values close to 0 for unseen areas, and values close to 1 for well-observed regions. We binarize the alpha channel and use this to filter out false changes produced from unseen areas. This produces our final multi-view change masks as follows:
\begin{equation}
\label{eq:intersec}
    M^k = M_{\text{ren}}^k \cdot \mathbf{1}(A_{\text{ren}}^k \geq 0.5).
\end{equation}

\subsection{Data Augmentation for Learning Change Channels}
\label{sec:method_dataaugment}
In this section, we explain how the set of individual image change masks can be augmented by also considering the \emph{reference} scene poses with a 3D representation of the \emph{inference} scene -- effectively reversing the change comparison between the scenes.

Following our pipeline, we obtain a change-aware 3DGS representing the inference scene, $\text{Change-3DGS}_{\text{inf}}$. This $\text{Change-3DGS}_{\text{inf}}$ can then be used to render inference scene (post-change) images for all reference scene (pre-change) viewpoints \( \mathcal{P}_{\text{ref}} \). Following the process outlined in Sec.~\ref{sec:method_changemasks}, we can then generate feature and structure-aware change masks by comparing the original \( \mathcal{I}_{\text{ref}} \) with these newly rendered images. These change masks can be concatenated with those initially calculated from the inference scene viewpoints \( \mathcal{P}_{\text{inf}} \) to create an augmented set of masks and once again re-optimize the change channels in $\text{Change-3DGS}_{\text{inf}}$ as described in Sec.~\ref{sec:method_changechannels} (see Supp. Material for a visualization).

\section{Experimental Setup}
\label{sec:exp_setup}

\subsection{Datasets}

We introduce the \textbf{Pose-Agnostic Scene-Level Change Detection Dataset (PASLCD)}, comprising data collected from ten complex, real-world scenes, including five indoor and five outdoor environments. PASLCD enables the evaluation of scene-level change detection, with multiple simultaneous changes per scene and ``distractor'' visual changes (i.e.~varying lighting, shadows, or reflections), 
% (see Fig.~\ref{fig:results_visualization} for a visualization of the scenes).
Among the indoor and outdoor scenes, two are 360\textdegree\space scenes, while the remaining three are front-facing (FF) scenes.

For all ten scenes in PASLCD, there are two available change detection instances: (1) change detection under consistent lighting conditions, and (2) change detection under varied lighting conditions. Images were captured using an iPhone following a random and independent trajectory for each scene instance. We provide 50 human-annotated change segmentation masks per scene, totaling 500 annotated masks for the dataset. Annotations were completed by two individuals following identical protocol -- rendering pre-change and post-change viewpoints, selecting the optimal viewpoint for change visibility and then using the Supervisely tool~\cite{supervisely} to annotate pixel-wise change masks.

Every inference scene contains multiple changes (between 5 to 17), encompassing both surface-level and structural changes. From a total of 91 changes across ten scenes, 70\% are structural changes, involving objects with 3D geometry being added (24\%), removed (27\%) or moved (18\%), with a range of object sizes and volumes from small and thin (e.g.~cutlery) to large and bulky (e.g.~benches), as well as challenging transparent glass objects. The other 30\% of changes are surface-level and involve minimal effect on the scene's 3D geometry, by adding or removing liquid spills and stickers (19\%) or changing surface colors (e.g.~swapping in structurally identical objects of different colors) (12\%). 
For a detailed description of the PASLCD dataset, we kindly refer readers to the Supp. Material. 

Additionally, we evaluate our method on simulated scene-level change detection dataset ChangeSim~\cite{park2021changesim} and the released subset of object-centric, pose-agnostic anomaly detection dataset MAD-Real~\cite{zhou2024pad}. 

\subsection{Baselines and Metrics}
\label{sec:metrics}
As a baseline, we test the ``Feature Difference'' (Feature Diff.) using our feature-aware change masks \( D^k_{\text{normalized}} \) calculated in Sec.~\ref{sec:method_changemasks}. This represents our method's performance before the inclusion of our key contributions with only per-view feature difference from a pre-trained model. We evaluate against two state-of-the-art approaches in pose-agnostic, self-supervised anomaly detection: OmniPoseAD~\cite{zhou2024pad} and SplatPose~\cite{kruse2024splatpose}. We also compare with supervised pairwise scene-level change detection (SCD) methods~\cite{sachdeva_change_2023,sachdeva_change_2023-1,sakurada_weakly_2020,varghese_changenet_2019} on the PASLCD and ChangeSim~\cite{park2021changesim} datasets. As these SCD methods are supervised, we use models pre-trained on COCO-Inpainted~\cite{sachdeva_change_2023-1} for CYWS-2D~\cite{sachdeva_change_2023-1} and CYWS-3D~\cite{sachdeva_change_2023}, and models pre-trained on ChangeSim~\cite{park2021changesim} for ChangeNet~\cite{varghese_changenet_2019} and CSCDNet~\cite{sakurada_weakly_2020}. CYWS-2D and 3D~\cite{sachdeva_change_2023,sachdeva_change_2023-1} predict change as bounding boxes, which we convert into a binary segmentation mask considering the area inside the bounding box. All SCD methods are evaluated on the aligned image pairs rendered from the reference scene, consistent with our Feature Diff. baseline.

Following the SCD literature \cite{sakurada_weakly_2020, sakurada_dense_2017, alcantarilla_street-view_2018, sakurada_change_2015, lei_hierarchical_2021, lin_robust_2024}, our primary evaluation metrics are mean Intersection over Union (mIoU) and F1 score, computed for ``change'' pixels in the ground-truth mask. 
% We compute these with respect to changed pixels in the ground truth masks, as our focus is on localizing changes. 
For MAD-Real~\cite{zhou2024pad}, we follow the initial evaluation and additionally report the Area Under the Receiver Operating Characteristic Curve (AUROC).

When calculating mIoU and F1, all methods are required to produce a scoreless binary mask (change vs. no change). Given that we are operating in a self-supervised setting without labels or a validation set, it is not possible to optimize for a threshold to convert continuous change masks into a binary mask. For all methods, we therefore threshold change masks with a value of 0.5 to provide a binarized change mask. We select 0.5 as it is the midpoint of possible change values, ranging from 0 to 1.

\section{Experimental Results}
\label{sec:results}

\subsection{Multi-view Pose-agnostic Change Localization}

\begin{table}[t]
    \centering
    \caption{Quantitative results for the MAD-Real~\cite{zhou2024pad} dataset, with results averaged over all ten LEGO object scenes.}
    \label{tab:results_mad}
    \vspace{-3pt}
    \scalebox{0.9}{ % Adjust the scaling factor as needed
        \begin{tabular}{lccc} \toprule
            Method & mIoU $\uparrow$ & F1 $\uparrow$ & AUROC $\uparrow$ \\ \midrule
            OmniPoseAD~\cite{zhou2024pad} & 0.064 & 0.115 & 0.937 \\ 
            SplatPose~\cite{kruse2024splatpose} & 0.077 & 0.123 & 0.898 \\
            Feature Difference & 0.052 & 0.089 & \textbf{0.967} \\    
            Ours & \textbf{0.132} & \textbf{0.210} & 0.953 \\ \bottomrule
            % \vspace{-25pt}
        \end{tabular}
        
    }
\end{table}

\noindent \textbf{Performance on Single-Object Scenes:} As shown in Tab.~\ref{tab:results_mad}, our method surpasses the state-of-the-art on the MAD-Real~\cite{zhou2024pad} single-object LEGO scenes. In particular, our mIoU achieves approximately a \( \textbf{1.7}\times \) improvement over SplatPose~\cite{kruse2024splatpose}, our closest competitor. 

\noindent \textbf{Performance on Multi-Object, Multi-Change Scenes: } In Tab.~\ref{tab:results_changesim}, we report results on simulated indoor industrial scenes from ChangeSim~\cite{park2021changesim} for Change (C) and Static (S) classes (following the established evaluation protocol). Our approach outperforms ChangeNet~\cite{varghese_changenet_2019} and CSCDNet~\cite{sakurada_weakly_2020}, with a \(\textbf{1.7}\times\) improvement in mIoU in the Change class.

\begin{table}[t]
    \centering
    \caption{Quantitative results for the ChangeSim~\cite{park2021changesim} dataset, with results averaged over test sequences. $^{*}$ results are taken from \cite{park2021changesim}.}
    \label{tab:results_changesim}
    \vspace{-3pt}
    \scalebox{0.9}{ % Adjust the scaling factor as needed
        \begin{tabular}{lcc} \toprule
            Method & Change (C) $\uparrow$ & Static (S) $\uparrow$ \\ \midrule
            ChangeNet~\cite{varghese_changenet_2019}$^{*}$ & 0.176 & 0.733  \\ 
            CSCDNet~\cite{sakurada_weakly_2020}$^{*}$ & 0.229 & 0.873 \\% & 0.207 & 0.895 \\ F1: 0.310 %& 0.215 & 0.873  \\ % 0.229 & xxx \\
            Feature Diff. & 0.352 & 0.807  \\ 
            Ours & \textbf{0.407} & \textbf{0.918}  \\ \bottomrule
            % \vspace{-25pt}
        \end{tabular}
    }
\end{table}

\begin{table*}[t]
    \centering
    \caption{Quantitative results for our dataset, averaged across similar and different lighting condition instances of both \colorbox[HTML]{EAEAEA}{Indoor} and \colorbox[HTML]{F0F8FF}{Outdoor} scenes. See Supp. Material for instance-level results. Our method consistently improves over the baselines in all instances.}
    \label{tab:results_main}
    \scalebox{0.83}{
        \begin{tabular}{lccccccccccccc}
            \toprule
            \multirow{2}{*}{Scene} & \multirow{2}{*}{FF/360} & \multicolumn{2}{c}{OmniPoseAD~\cite{zhou2024pad}} & \multicolumn{2}{c}{SplatPose~\cite{kruse2024splatpose}} & \multicolumn{2}{c}{CSCDNet~\cite{sakurada_weakly_2020}} & \multicolumn{2}{c}{CYWS-2D~\cite{sachdeva_change_2023-1}} & \multicolumn{2}{c}{Feature Diff.} & \multicolumn{2}{c}{Ours} \\
            \cmidrule(lr){3-4} \cmidrule(lr){5-6} \cmidrule(lr){7-8} \cmidrule(lr){9-10} \cmidrule(lr){11-12} \cmidrule(lr){13-14}
            & &  mIoU $\uparrow$ & F1 $\uparrow$ & mIoU $\uparrow$ & F1 $\uparrow$ & mIoU $\uparrow$ & F1 $\uparrow$ & mIoU $\uparrow$ & F1 $\uparrow$ & mIoU $\uparrow$ & F1 $\uparrow$ & mIoU $\uparrow$ & F1 $\uparrow$ \\
            \midrule
            \rowcolor[HTML]{EAEAEA} Cantina & FF & 0.138 & 0.231 & 0.188 & 0.304 & 0.079 & 0.138 & 0.277 & 0.408 & 0.251 & 0.382 & \textbf{0.580} & \textbf{0.729}\\
            \rowcolor[HTML]{EAEAEA} Lounge & FF & 0.149 & 0.241 & 0.262 & 0.410 & 0.195 & 0.318 & 0.221 & 0.348 & 0.177 & 0.296 & \textbf{0.463} & \textbf{0.626} \\
            \rowcolor[HTML]{EAEAEA} Printing Area & FF & 0.157 & 0.242 & 0.183 & 0.288 & 0.147 & 0.246 & 0.327 & 0.457 & 0.432 & 0.584 & \textbf{0.588} & \textbf{0.734}\\
            \rowcolor[HTML]{EAEAEA} Lunch Room & 360 & 0.161 & 0.247 & 0.133 & 0.215 & 0.035 & 0.062 & 0.123 & 0.204 & 0.101 & 0.174 & \textbf{0.389} & \textbf{0.546} \\
            \rowcolor[HTML]{EAEAEA} Meeting Room & 360 & 0.107 & 0.182 & 0.130 & 0.211 & 0.213 & 0.330 & 0.138 & 0.233 & 0.122 & 0.211 & \textbf{0.350} & \textbf{0.507} \\
            \midrule
            \rowcolor[HTML]{F0F8FF} Garden & FF & 0.273 & 0.411 & 0.185 & 0.300 & 0.241 & 0.383 & 0.346 & 0.509 & 0.292 & 0.445 & \textbf{0.436} & \textbf{0.601} \\
            \rowcolor[HTML]{F0F8FF} Pots & FF & 0.143 & 0.230 & 0.140 & 0.290 & 0.022 & 0.041 & 0.351 & 0.531 & 0.397 & 0.566 & \textbf{0.540} & \textbf{0.693} \\
            \rowcolor[HTML]{F0F8FF} Zen & FF & 0.205 & 0.317 & 0.186 & 0.296 & 0.010 & 0.018 & 0.450 & 0.575 & 0.444 & 0.577 & \textbf{0.500} & \textbf{0.633}  \\
            \rowcolor[HTML]{F0F8FF} Playground & 360 & 0.076 & 0.125 & 0.081 & 0.133 & 0.134 & 0.219 & 0.059 & 0.107 & 0.047 & 0.089 & \textbf{0.249} & \textbf{0.378}\\
            \rowcolor[HTML]{F0F8FF} Porch & 360 & 0.274 & 0.396 & 0.239 & 0.363 & 0.176 & 0.292 & 0.439 & 0.607 & 0.379 & 0.538 & \textbf{0.518} & \textbf{0.676} \\
            \midrule
            Average & -- & 0.168 & 0.262 & 0.173 & 0.281 & 0.125 & 0.204 & 0.273 & 0.398 & 0.264 & 0.386 & \textbf{0.461} & \textbf{0.612} \\
            \bottomrule
        \end{tabular}
    }
\end{table*}

In Tab.~\ref{tab:results_main}, we present results for each scene in our PASLCD dataset averaged across the two instances with varying lighting conditions. Our method consistently outperforms all baselines, validating our claim that we achieve state-of-the-art performance for multi-object scene change detection -- we achieve approximately \( \textbf{1.7}\times \) improvement in mIoU and \( \textbf{1.5}\times \) in F1 score over the best competitor.

\begin{figure*}[t]
    \centering
    \includegraphics[width=\textwidth]{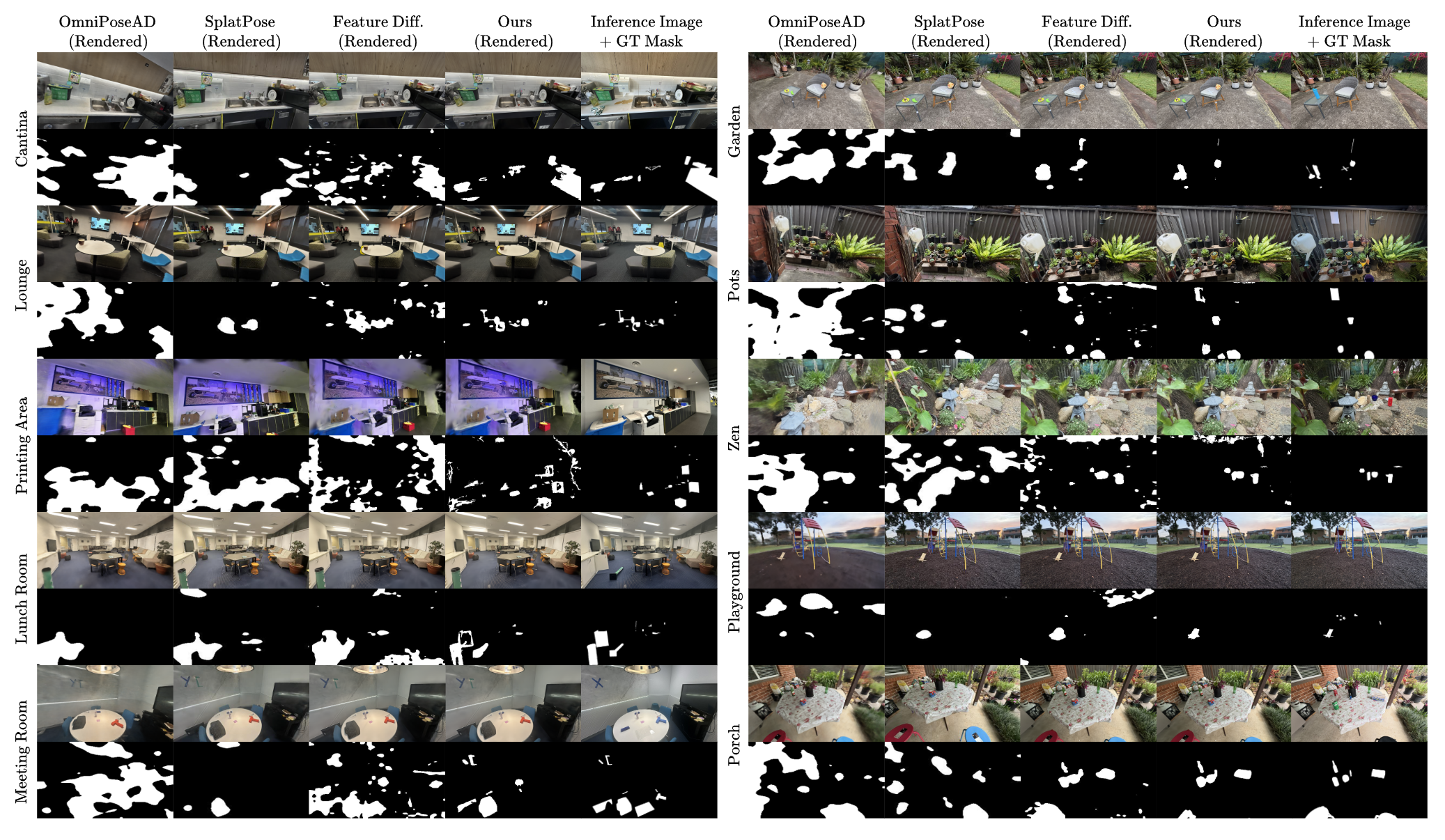}
    \vspace{-20pt}
     \caption{Qualitative results of each approach on our PASLCD dataset. See Supp. Material for additional visualizations. Our generated change masks consistently agree more closely with the ground truth compared to the baselines.}
     \vspace{-0.2cm}
    \label{fig:results_visualization}
\end{figure*}

\noindent\textbf{Qualitative Results:} Fig.~\ref{fig:results_visualization} presents a randomly sampled example change detection from each scene for all methods. Prior state-of-the-art methods OmniPoseAD~\cite{zhou2024pad} and SplatPose~\cite{kruse2024splatpose} scale poorly to multi-object scenes, with the optimization-based pose estimation often failing to converge to a global minimum (see the Cantina, Printing Area, and Pots scenes in Fig.~\ref{fig:results_visualization}). Convergence frequently fails when inference images lack sufficient overlap with the images in the reference set and the methods cannot obtain a reasonable coarse pose estimation for the optimization. 

We also observe some consistent failure cases of our multi-view change masks in Fig.~\ref{fig:results_visualization}: (1) identifying color-based surface-level changes (spill on the bench in Cantina scene and T block color change in Meeting Room scene). Upon investigation, this is due to the failure of the pre-trained foundation model to produce feature changes in these conditions; (2) difficulty identifying very small changes in large-scale scenes (see Playground and Lunch Room scenes); (3) overestimating change masks for true changes, due to the patch-to-pixel interpolation of our feature masks. This is observed to a greater degree in the Feature Difference baseline. In the Supp. Material, we also include visualizations highlighting different types of failure cases (false positive vs. false negative change predictions).

\subsection{Comparison with Pair-wise Scene Change Detection Approaches}
\label{sec:scd} 
In Tab.~\ref{tab:comp2d}, we compare the performance of SCD pair-wise change masks~\cite{sachdeva_change_2023,sachdeva_change_2023-1,sakurada_weakly_2020, varghese_changenet_2019} and our proposed pair-wise feature and structure-aware masks, showing that our proposed approach achieves best performance. In contrast to our approach, the other SCD baselines~\cite{sachdeva_change_2023,sachdeva_change_2023-1,sakurada_weakly_2020, varghese_changenet_2019} use supervised learning to generate change masks, assuming training on large-scale datasets matching the test-time change distribution. Performance can suffer (see Tab.~\ref{tab:results_main}) when this is violated, e.g.~CSCDNet trained on ChangeSim (simulated indoor industrial scenes) and testing on PASLCD (real generic indoor and outdoor scenes). 

Importantly, our Change-3DGS can be used as a multi-view extension for \emph{any} existing method of change mask generation, boosting performance by enforcing multi-view consistency. In Tab.~\ref{tab:comp2d}, we show that the mIoU of CYWS-2D~\cite{sachdeva_change_2023-1} increases by 44\% when combined with our Change-3DGS to enable multi-view change consistency.

\begin{table}[t]
    \centering
    \caption{Quantitative results for pair-wise scene change detection baselines on PASLCD (averaged over all 20 instances).}
    \label{tab:comp2d}
    \vspace{-3pt}
    \scalebox{0.85}{ % Adjust the scaling factor as needed
        \begin{tabular}{lcccc} \toprule
            Method & mIoU $\uparrow$ & F1 $\uparrow$ 
            \\ \midrule
            CSCDNet~\cite{sakurada_weakly_2020}  & 0.125 & 0.204 \\
        CYWS-3D~\cite{sachdeva_change_2023}  & 0.201  & 0.303 \\ 
        CYWS-2D~\cite{sachdeva_change_2023-1}   & 0.273 & 0.388\\
        % Feature Diff. & 0.264 & 0.386 \\
        Feature \& Structure-Aware Mask \textbf{(Ours)} & 0.372 & 0.519 \\
        \midrule
        CYWS-2D~\cite{sachdeva_change_2023-1} + Change-3DGS (\textbf{Ours}) & 0.393 & 0.538 \\
         F\&S Mask (\textbf{Ours}) + Change-3DGS (\textbf{Ours}) & \textbf{0.449} & \textbf{0.598} \\
        \bottomrule
        \end{tabular}
    } 
\end{table}

\subsection{Performance with Limited Inference Views}
\label{sec:n_views}
\begin{figure}
    \centering
    \includegraphics[width=1\linewidth]{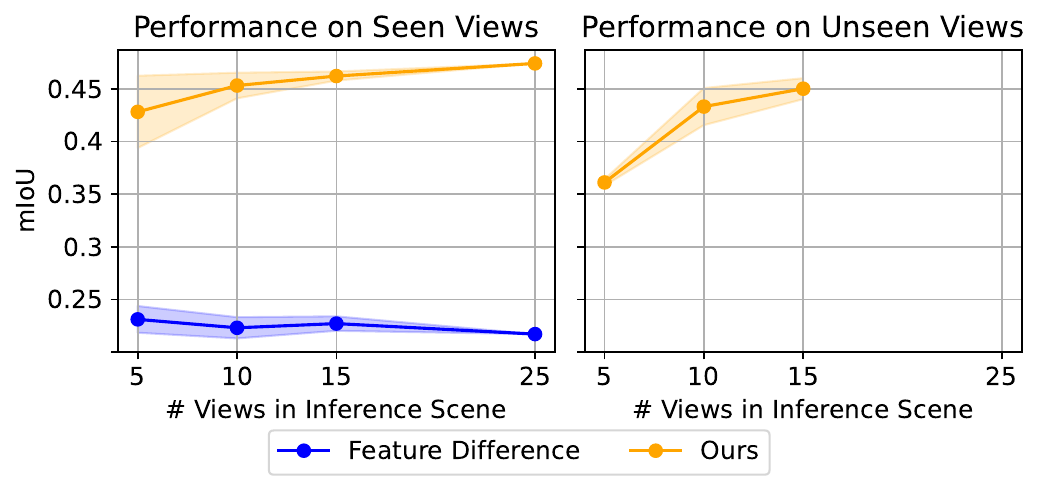}
    % \vspace{-20pt}
    \caption{Performance with varying numbers of inference views.}
    \label{fig:results_multiview}
    \vspace{-0.3cm}
\end{figure}

In Fig.~\ref{fig:results_multiview}, we explore how the number of images observed in the inference scene ($n_{\text{inf}}$) influences the performance of our multi-view change masks on seen and unseen views. For all indoor scenes in our PASLCD dataset, we randomly sample 5, 10, and 15 images for seen views from the total 25 available images in the scene. We hold-out poses of 10 images from the remaining 25 images as unseen views. We report the mean and standard deviation across 3 random trials.

\noindent\textbf{Robustness to Limited Inference Scene Views: } As shown in the left-hand of Fig.~\ref{fig:results_multiview}, our method's performance increases as more views from the inference scene can be leveraged for the multi-view change masks. Notably, even with only 5 images from the inference scene, our method is still able to outperform the Feature Diff. baseline by an impressive margin (approximately 1.8$\times$ the mIoU) averaged over all trials. As expected, the Feature Diff. baseline maintains consistent performance regardless of the number of images in the inference scene, as it treats images individually when generating change masks.

\noindent\textbf{Generating Change Masks for Unseen Views:} We also validate our claim that our method can generalize to unseen views by generating change masks for query poses that \emph{have not been observed} in the inference scene. This is a new capability unlocked by our change detection method that has not been previously explored -- only by embedding change information in a 3D representation can we render change masks for entirely unseen views.

For each trial, we render change masks for the 10 unseen query poses (there are only 25 images per scene so there are no unseen views when using 25 inference views). The right-hand of Fig.~\ref{fig:results_multiview} shows that our approach generates change masks for \emph{unseen} views that outperform the Feature Diff. baseline on \emph{seen} data, with mIoU ranging between 0.36-0.45 on average depending on the number of views in the inference scene used to learn the multi-view change masks.

\subsection{Ablations}
\label{sec:ablation_sh}
\textbf{Spherical Harmonics Degree:} 
Tab.~\ref{tab:ablation_sh} validates our hypothesis that lower degrees of spherical harmonics (SH) coefficient allow the 3DGS to remove view-dependent false positive change predictions. Results are averaged over the 5 indoor scenes in our PASLCD dataset and show that the lowest SH degree provides the best mIoU and F1 for our multi-view change masks. We also report the average number of false positive (FP) and false negative (FN) pixels per image, showing an approximate $70\%$ reduction in false change predictions (FPs) between the highest and lowest SH degrees. As expected, inhibiting view-dependent change modeling with lower SH degrees also introduces a slight trade-off with increased missed changes (FNs), although not outweighing the gains from reduced FPs.

\begin{table}[tb]
    \centering
    \caption{Quantitative results for varying SH degree. Lower SH degrees yield better change detection performance.}
    \label{tab:ablation_sh}
    \vspace{-3pt}
    \scalebox{0.9}{ % Adjust the scaling factor as needed
        \begin{tabular}{ccccc} \toprule
            SH Degree & mIoU $\uparrow$ & F1 $\uparrow$ & \# $\text{FP}_{\text{im}}$ $\downarrow$ & \# $\text{FN}_{\text{im}}$ $\downarrow$ \\ \midrule
            0 & \textbf{0.4741} & \textbf{0.628} & \textbf{879} & 534 \\
            1 &  0.460 & 0.614 & 1030 & 478 \\
            2 & 0.453 & 0.607 & 1142 & 442 \\
            3 & 0.442 & 0.596 & 1257 & \textbf{416}\\ \bottomrule
        \end{tabular}
    }
\end{table}

\noindent\textbf{Ablation on Different Modules:} 
Tab.~\ref{tab:ablation_modules} shows the performance contributed by the individual modules in our proposed method: (1) the Feature Difference baseline, (2) learning a Change-3DGS with only feature-aware masks, (3) a Change-3DGS with only structure-aware masks, (4) our proposed Change-3DGS, (4) when including data augmentation, and (5) when accounting for unseen regions with the alpha channel. In particular, we validate our claim that the feature-aware mask and structure-aware masks contain complementary information that can be combined for best performance -- their combined mIoU performance improves the performance of either alone by a factor of approximately $\textbf{1.4}\times$ (see further discussion in Supp. Material). 

\begin{table}[tb]
    \centering
    \caption{Ablation of our method reported on PASLCD.}
    \label{tab:ablation_modules}
    \vspace{-3pt}
    \scalebox{0.9}{ % Adjust the scaling factor as needed
        \begin{tabular}{lcc}\toprule
            Component & mIoU $\uparrow$ & F1 $\uparrow$ \\ \midrule
            Feature Difference   & 0.264 & 0.382  \\ 
            Change-3DGS ({\small feature-aware masks}) & 0.311 & 0.448 \\
            Change-3DGS ({\small structure-aware masks})  & 0.324 & 0.461 \\
            Change-3DGS ({\small combined}) & 0.449 & 0.598 \\
            Learned Mask + Aug.  & 0.457 & 0.605 \\ 
            Learned Mask + Aug. + Alpha Channel & \textbf{0.461} & \textbf{0.612}\\
            \bottomrule
        \end{tabular}
    }
\end{table}

\section{Conclusion}
We presented a new state-of-the-art multi-view approach to label-free, pose-agnostic change detection. We integrate multi-view change information into a 3DGS representation, enabling robust change localization even for unseen viewpoints. We additionally introduced a new change detection dataset featuring multi-object real-world scenes, which we hope will drive further advancements in the change detection community. Future work should focus on addressing the limitations observed in the feature masks from the foundation model, namely difficulty identifying surface-level changes and difficulty producing refined change masks.

{
    \small
    \bibliographystyle{ieeenat_fullname}
    \bibliography{main}
}

% WARNING: do not forget to delete the supplementary pages from your submission 
\clearpage
\setcounter{page}{1}
\maketitlesupplementary
\section{Additional Details on our Methodology}

\subsection{Motivation for change-specific opacity factor}
\label{sec:supp_change_motivation}
As discussed in Sec.~\ref{sec:method_changechannels}, our Change-3DGS can render both RGB images of the inference scene and change maps in parallel. To achieve this, we incorporate a separate opacity factor (\(\tilde{\alpha}\)) -- we explain the necessity of this design decision below.

During optimization, the standard 3DGS process~\cite{kerbl3Dgaussians} uses the opacity factor (\(\alpha\)) to identify when Gaussians do not contribute to the modeling and should be culled. In our change detection scenario, there can be situations where the Gaussians required to model RGB appearance versus change maps can differ. For example, consider scenarios where an object present in the reference scene is missing or has been moved in the inference scene. In the standard 3DGS process, Gaussians representing such missing/moved structures lower their opacity (\(\alpha\)) over the training as they are not visible in the set of inference images \(\mathcal{I}_{\text{inf}}\), eventually becoming transparent and being pruned. However, for change modeling, these Gaussians can be critical structures for embedding change in a change mask, carrying high change magnitudes (\(\tilde{c}\)). For this reason, we incorporate a separate change opacity factor into each Gaussian and consider both opacity factors (\(\alpha\) and \(\tilde{\alpha}\)) when determining whether a Gaussian should be removed, applying the minimum opacity threshold \(\epsilon_{\alpha}\)~\cite{kerbl3Dgaussians}. Gaussians are only removed when both \(\alpha\) and \(\tilde{\alpha}\) fall below the culling threshold.

\subsection{Motivation for initializing Change-3DGS with reference scene 3DGS}
We initialize our Change-3DGS with the existing 3DGS for the reference scene for two reasons: (1) many underlying structural elements of the scene are likely to remain consistent between the two scenes, and leveraging the already built reference 3DGS can allow us to update for an inference 3DGS with less data than learning from scratch; (2) as described in Sec.~\ref{sec:supp_change_motivation}, the reference scene can contain Gaussians representing structures that disappear in the inference scene and are important for modeling change -- these can be challenging to learn if learning the inference 3DGS from scratch.

\subsection{Visualization of Data Augmentation for Learning Change Channels}
\begin{figure}[!t]
    \centering
    \includegraphics[width=\linewidth,clip,trim={0 2cm 0 0}]{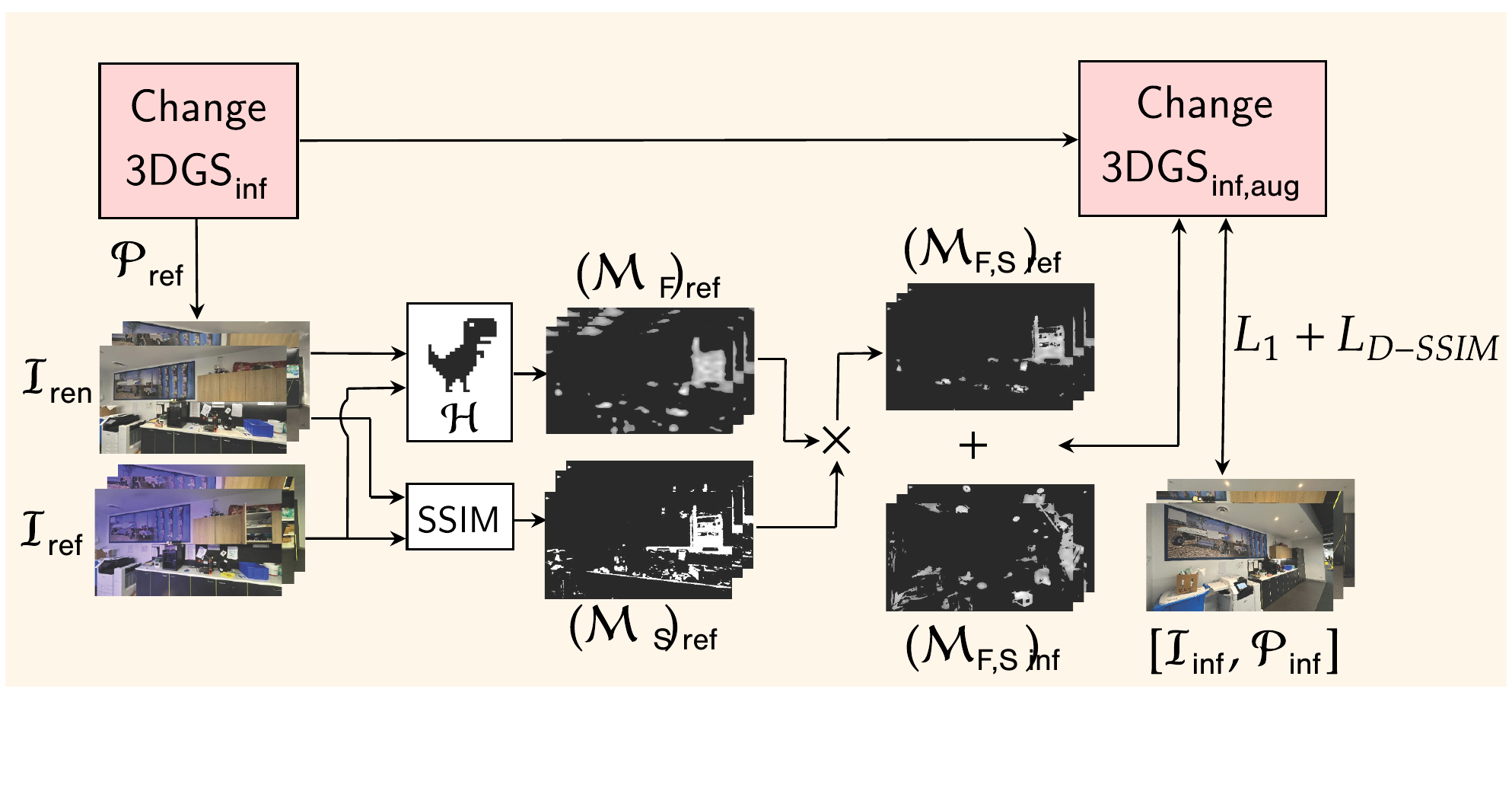}
    \caption{An overview of our data augmentation method. We concatenate the candidate masks $(\mathcal{M}_{\text{F,S}})_{\text{inf}}$ generated following Fig.~\ref{fig:overview} with candidate masks $(\mathcal{M}_{\text{F,S}})_{\text{ref}}$ obtained by considering the inference scene's representation viewed from the reference scene's poses.}
    \label{fig:supp_data_aug}
\end{figure}

We visualize the data augmentation process described in Sec.~~\ref{sec:method_dataaugment} in Fig.~\ref{fig:supp_data_aug}.

\subsection{Additional Implementation Details}
We build the reference scene by training on \(\mathcal{I}_{\text{ref}}\) and \(\mathcal{P}_{\text{ref}}\) for 7000 iterations. Once initiated with a reference scene, we only train for 3000 iterations to update the representation to inference scene with \(\mathcal{I}_{\text{inf}}\) and \(\mathcal{P}_{\text{inf}}\) while simultaneously optimizing the change channel guided by \(M_{F,P}\) (see Sec.~\ref{sec:method_changechannels}). Once the inference scene representation is built, we fine-tune the change channel for another 3000 iterations using the augmented candidate change mask following the process described in Sec.~~\ref{sec:method_dataaugment}. All the experiments were conducted on a single NVIDIA RTX 4090 GPU.

\section{Additional Details on Datasets}
\subsection{Additional Details on MAD-Real} \label{sec:supp_mad_real}

The MAD-Real dataset~\cite{zhou2024pad} has publicly released 10 scenes each containing a LEGO toy object. We illustrate each scene at the end of this Supp. Material: Bear, Bird, Elephant, Parrot, Pig, Puppy, Scorpion, Turtle, Unicorn, and Whale. 
% In Fig.~\ref{fig:supp_mad_real}, we visualize a few random images from each scene, taken from the train-set (faultless) and test-set (faulty) and the respective ground truth mask for the test image.
During our experiments, we consider the train-set as the image set for the reference scene and the test-set as the image set for the inference scene. 

\subsection{Additional Details on PASLCD}

We provide a breakdown of the change types and prevalence represented in PASLCD in Fig.~\ref{fig:supp_paslcd_stats}. A wide range of change prevalence is tested, ranging between 0.17\% and 20.12\%, with an average of 3.51\%.

Each figure contains a set of images from the inference scene, a set of images from the reference scene collected under similar lighting conditions to the inference images (Instance 1), and a set of images taken from the reference scene collected under different lighting conditions (Instance 2). The inference set is annotated with respect to Instance 1 and Instance 2. 

Images were captured using an iPhone with a 16:9 aspect ratio. For each instance, a human inspector independently moved across the scene following a random trajectory, while capturing the scene with no constraints on the camera pose. Images were taken at random heights and random orientations. 

\begin{figure*}[t]
    \centering
    \includegraphics[width=\textwidth]{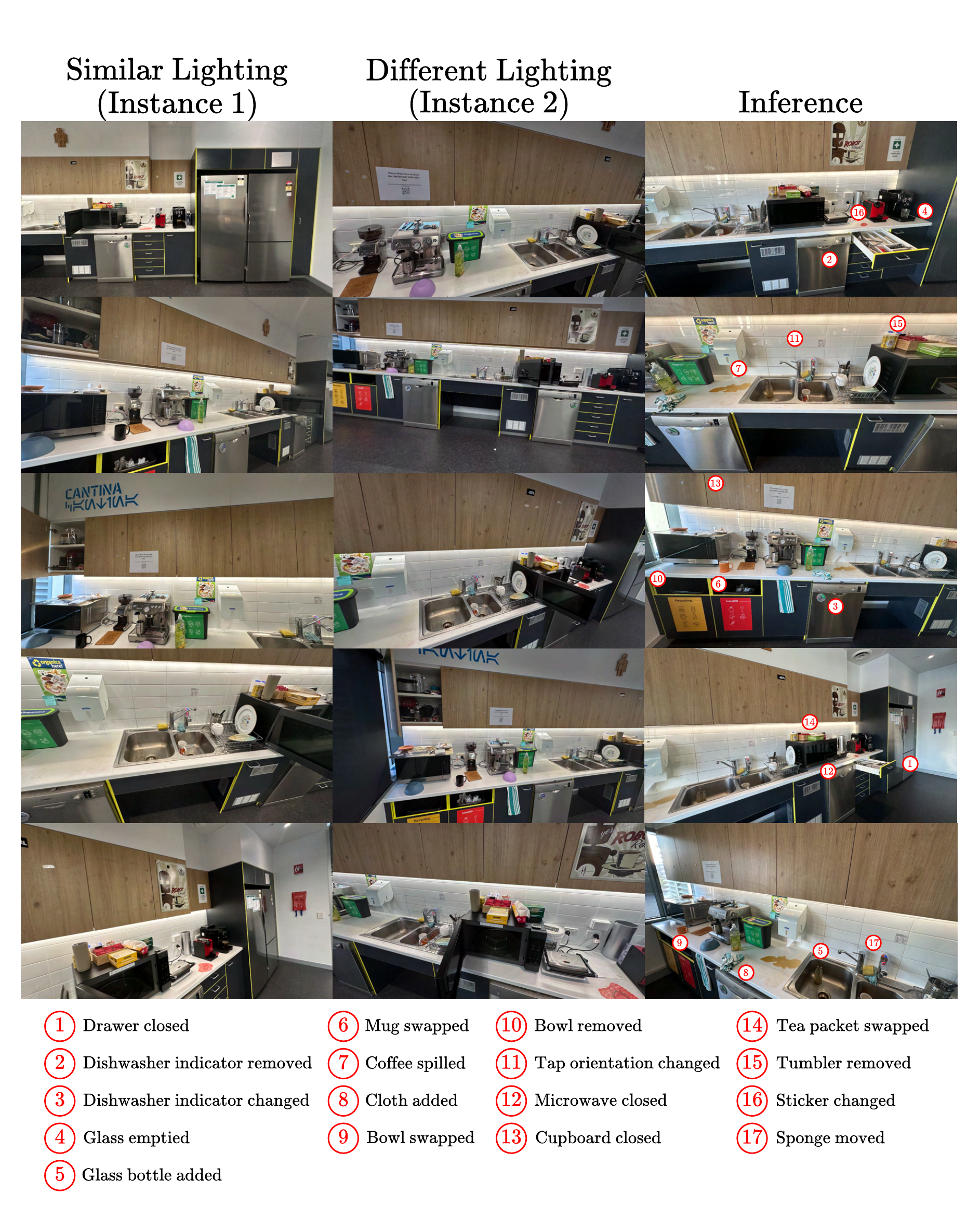}
    \caption{Cantina scene visualizations and change descriptions.}
    \label{fig:supp_cantina}
\end{figure*}

\begin{figure*}[t]
    \centering
    \includegraphics[width=\textwidth]{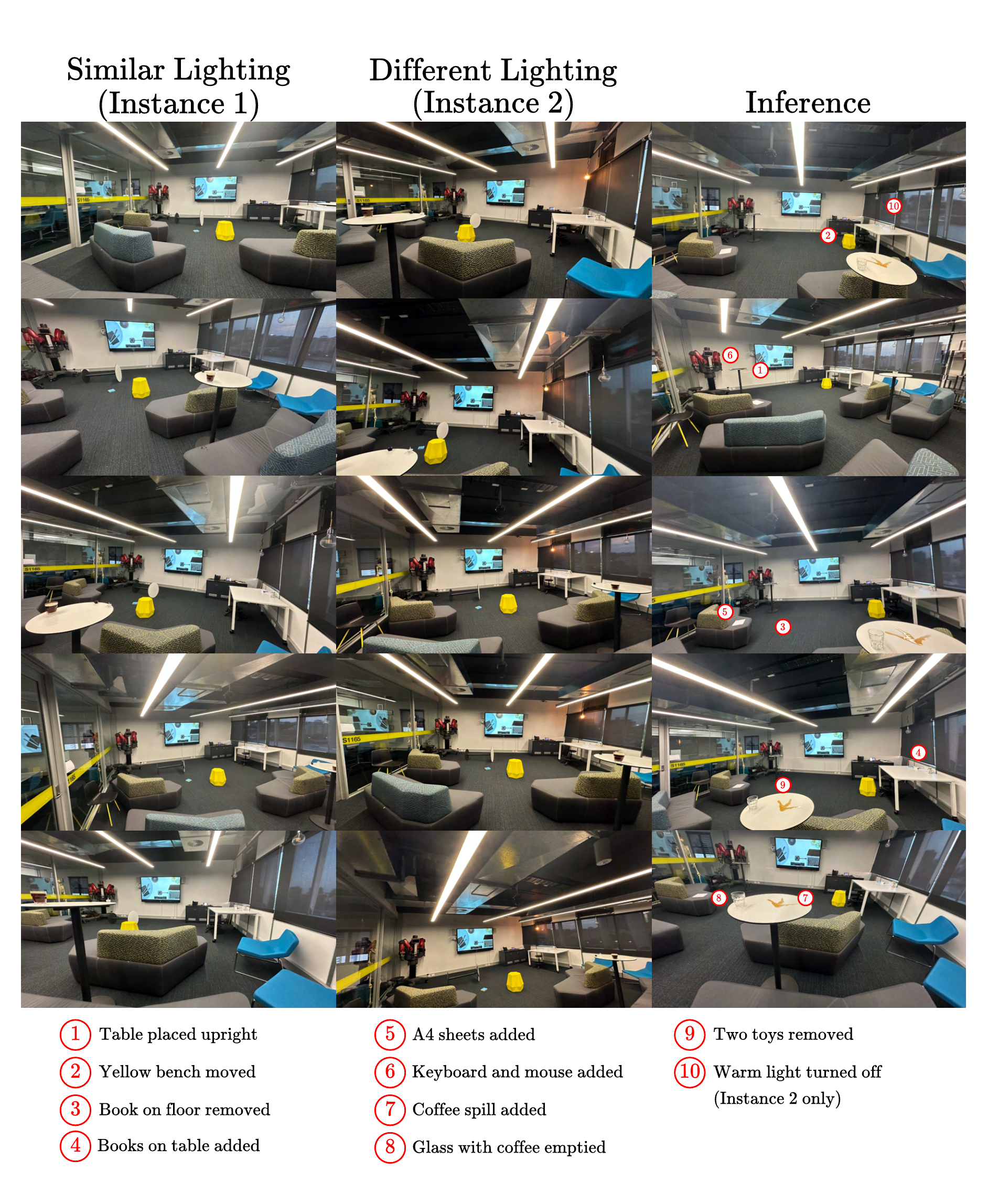}
    \caption{Lounge scene visualizations and change descriptions.}
    \label{fig:supp_lounge}
\end{figure*}

\begin{figure*}[t]
    \centering
    \includegraphics[width=\textwidth]{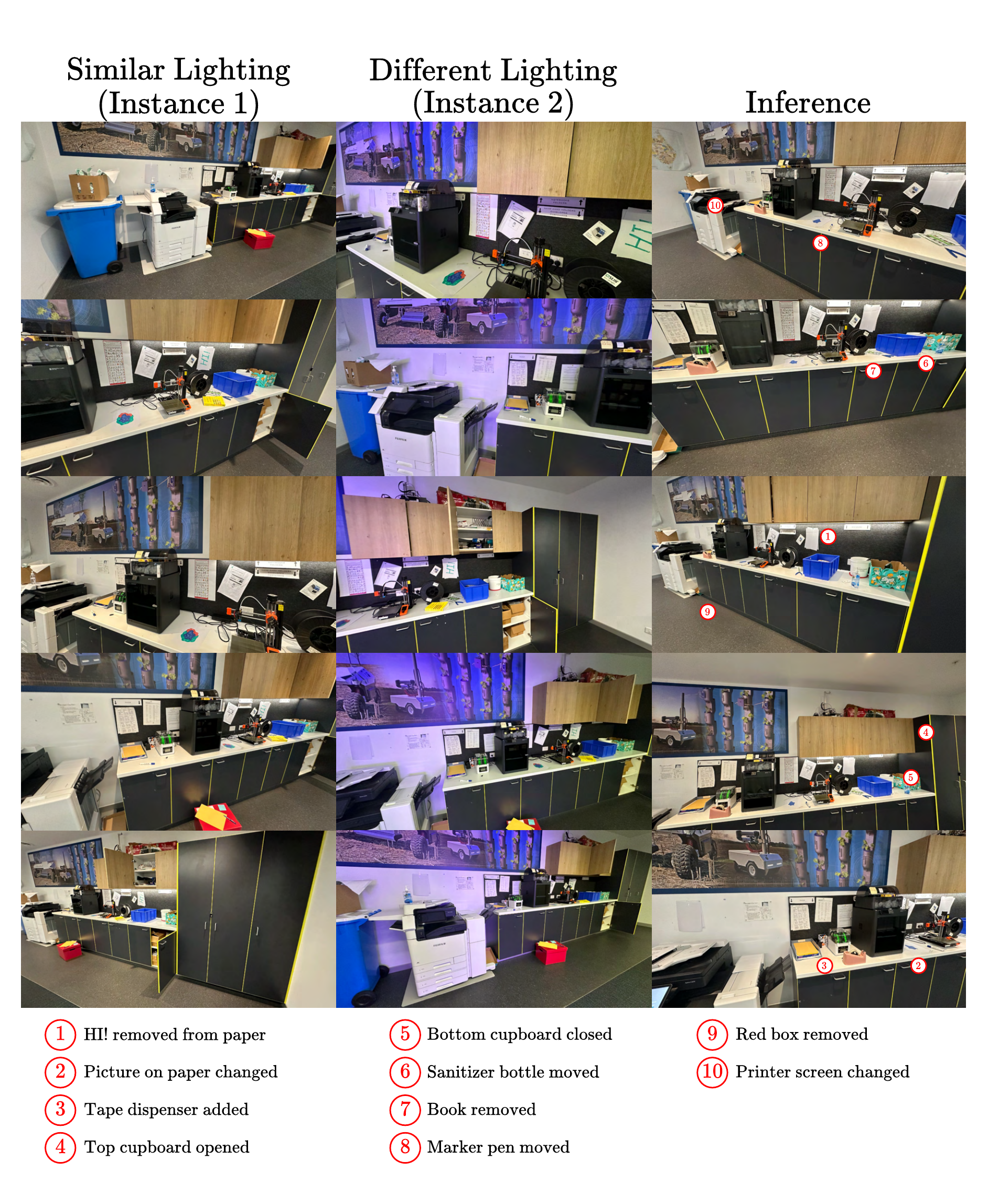}
    \caption{Printing area scene visualizations and change descriptions.}
    \label{fig:supp_printing}
\end{figure*}

\begin{figure*}[t]
    \centering
    \includegraphics[width=\textwidth]{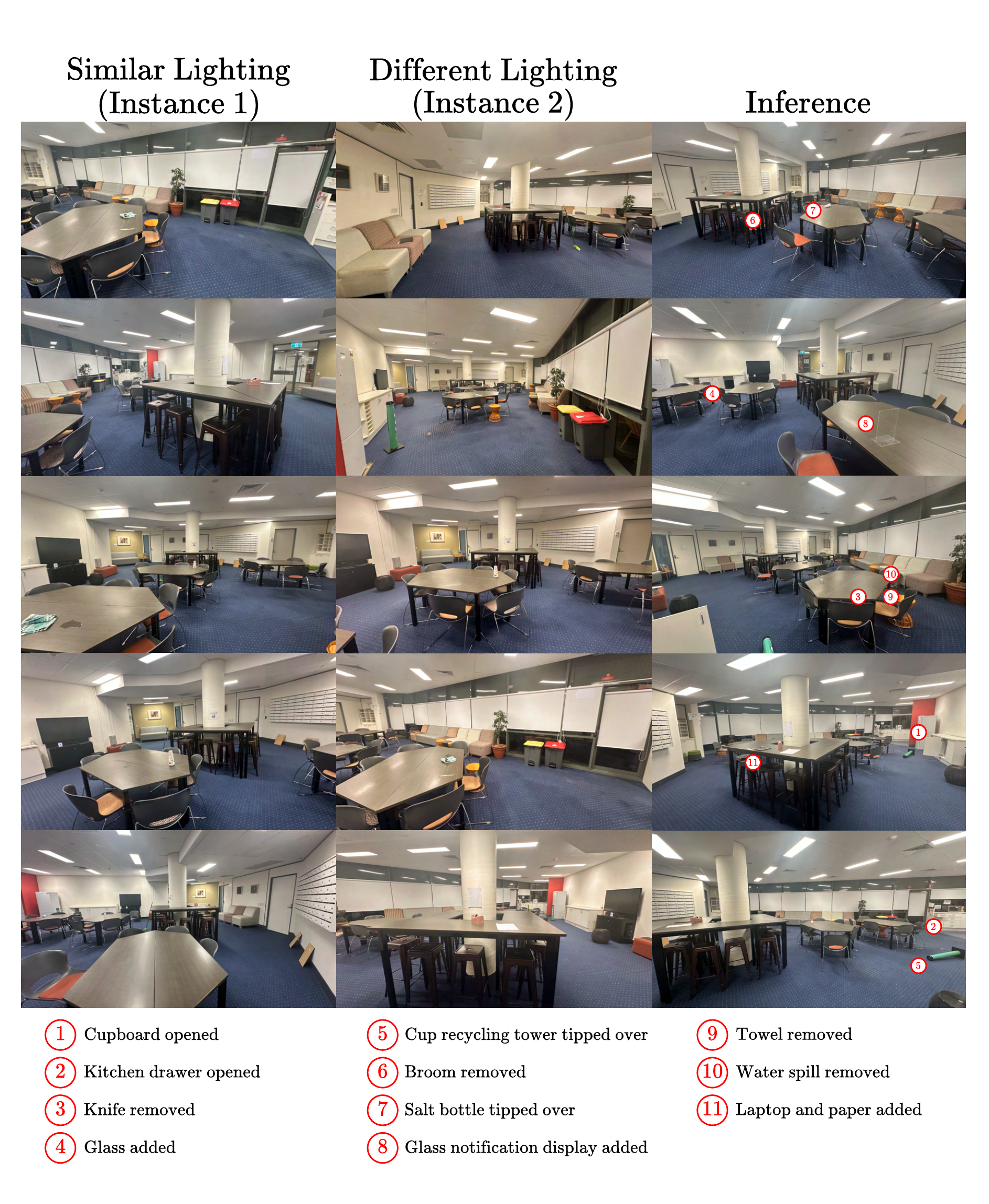}
    \caption{Lunch room scene visualizations and change descriptions.}
    \label{fig:supp_lunch}
\end{figure*}

\begin{figure*}[t]
    \centering
    \includegraphics[width=\textwidth]{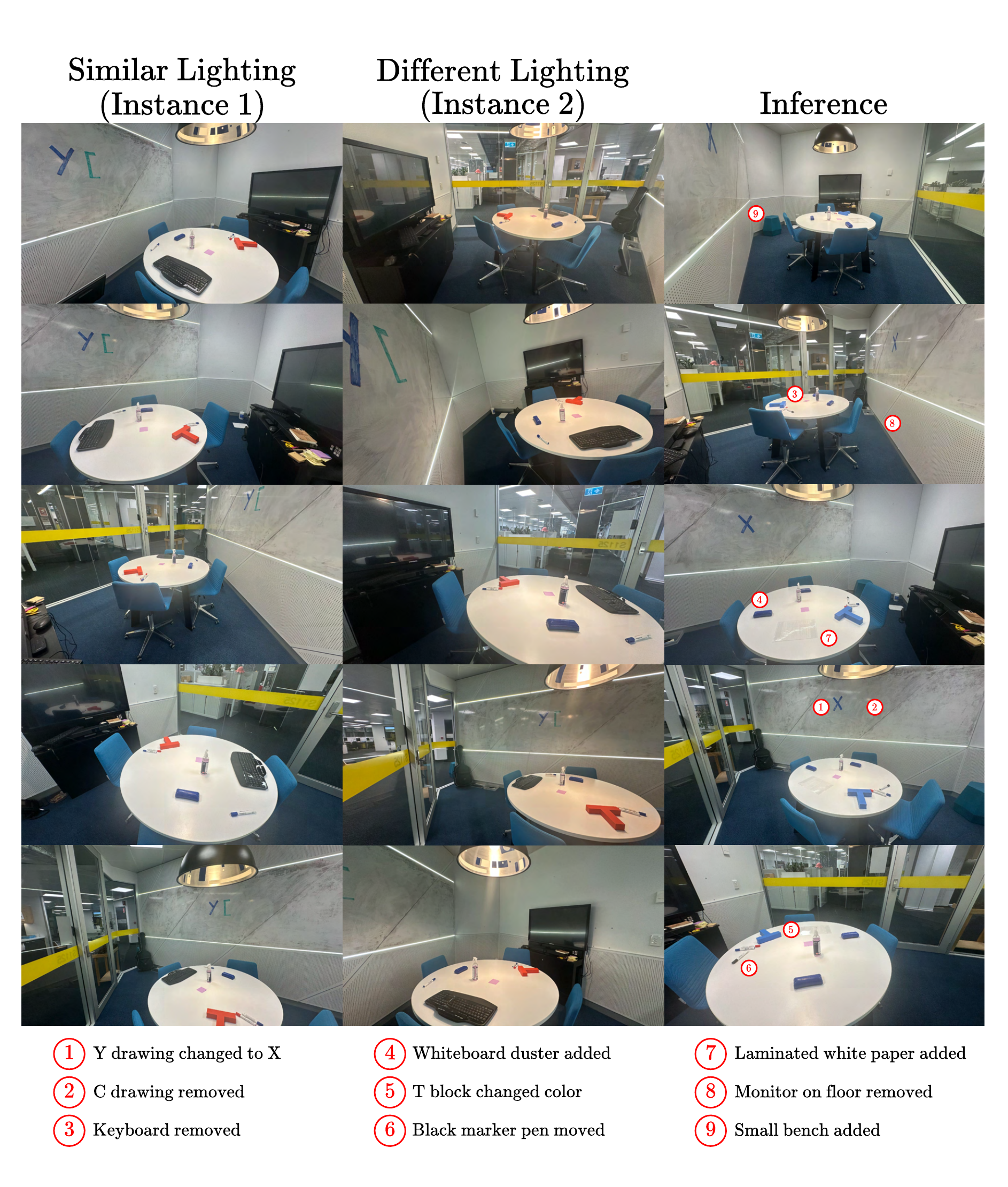}
    \caption{Meeting room scene visualizations and change descriptions.}
    \label{fig:supp_meeting}
\end{figure*}

\begin{figure*}[t]
    \centering
    \includegraphics[width=\textwidth]{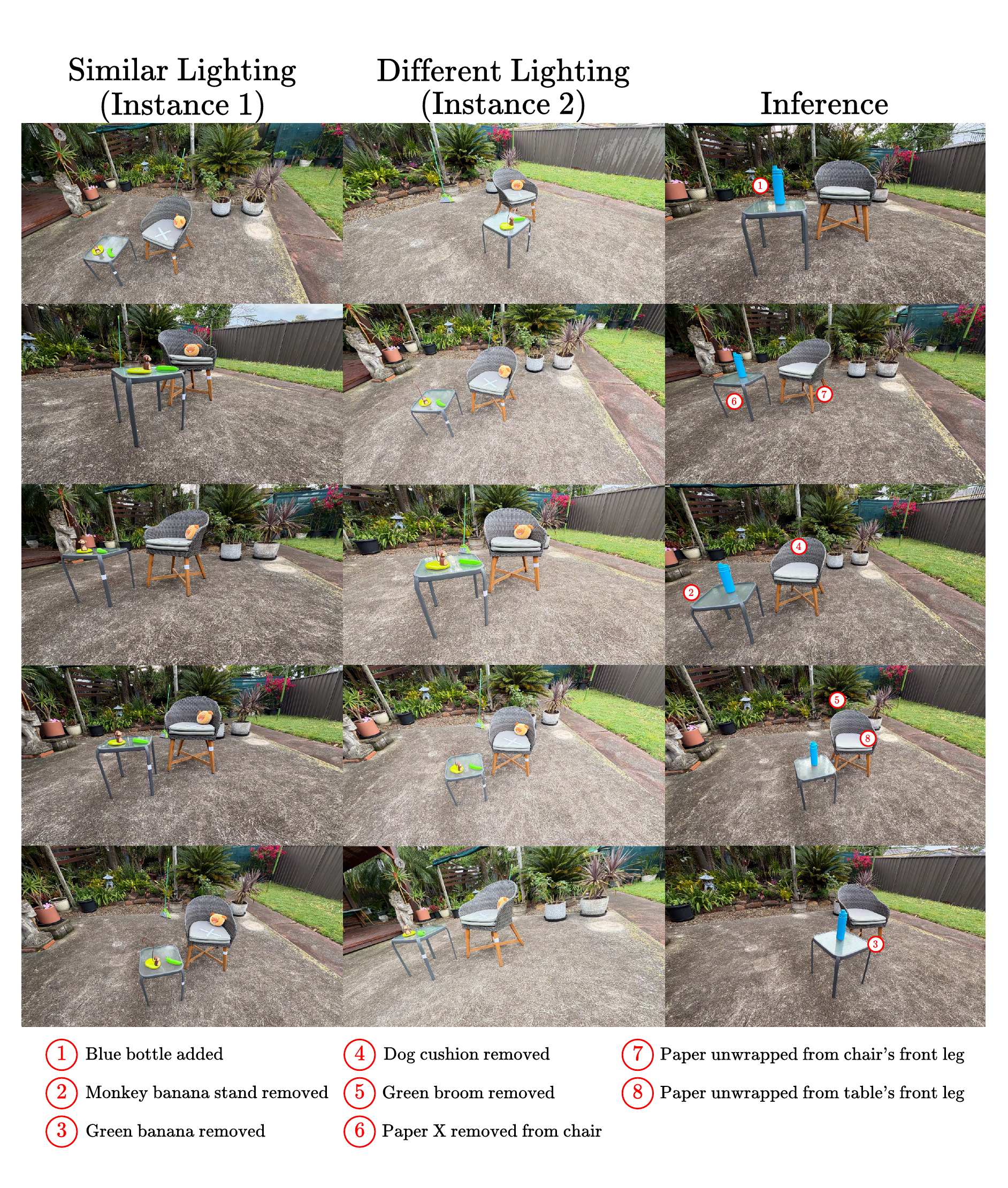}
    \caption{Garden scene visualizations and change descriptions.}
    \label{fig:supp_garden}
\end{figure*}

\begin{figure*}[t]
    \centering
    \includegraphics[width=\textwidth]{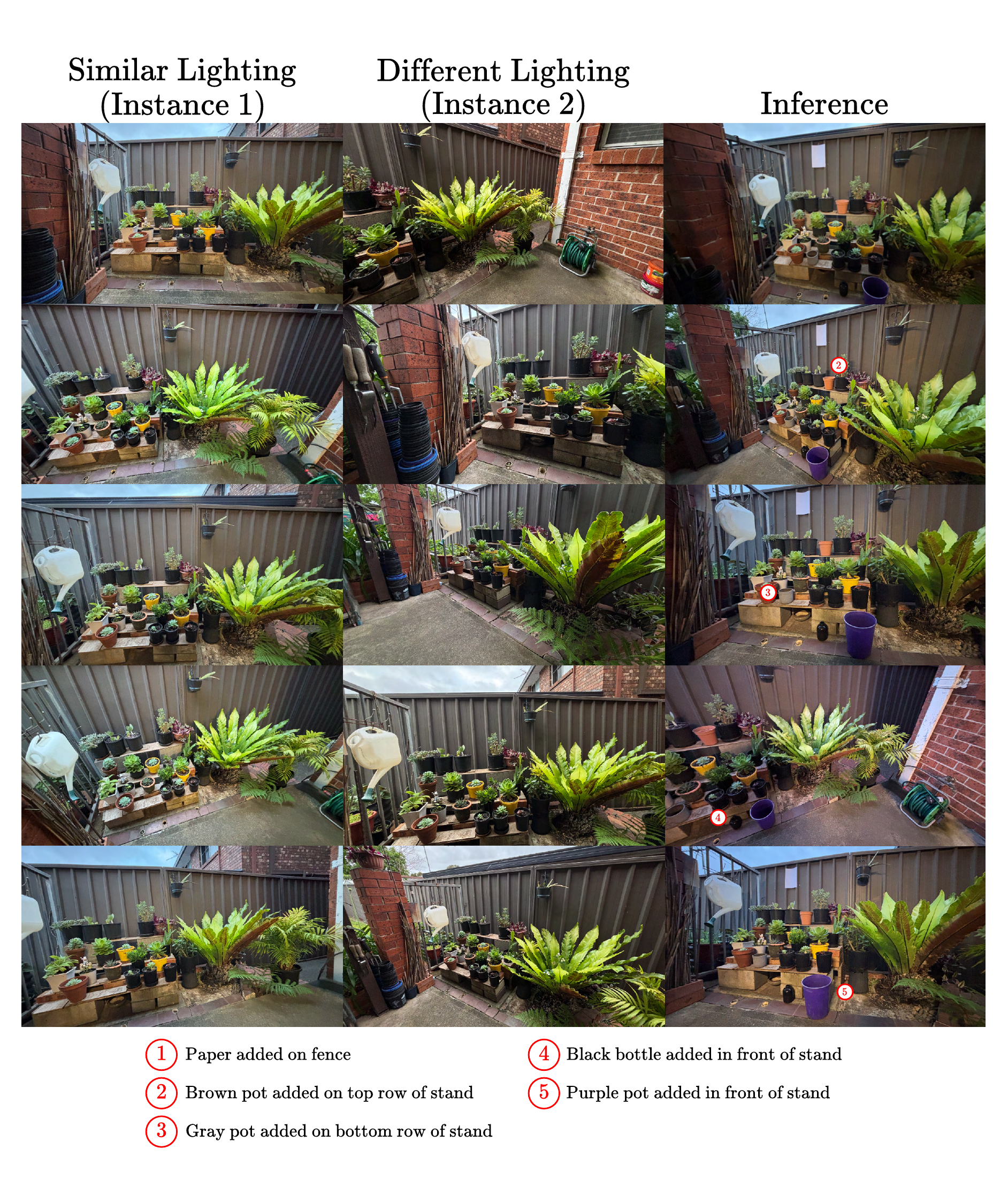}
    \caption{Pots scene visualizations and change descriptions.}
    \label{fig:supp_pots}
\end{figure*}

\begin{figure*}[t]
    \centering
    \includegraphics[width=\textwidth]{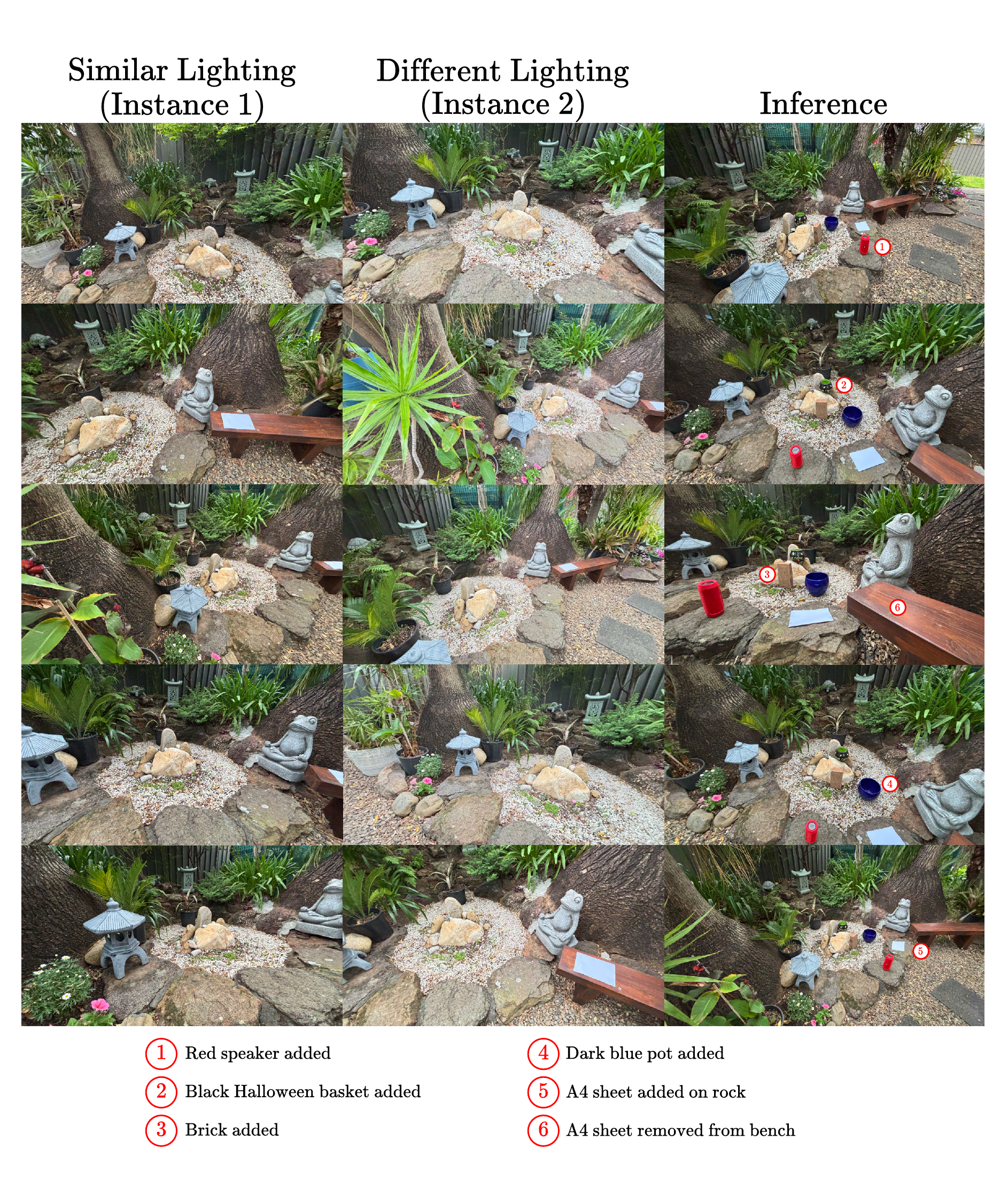}
    \caption{Zen scene visualizations and change descriptions.}
    \label{fig:supp_zen}
\end{figure*}

\begin{figure*}[t]
    \centering
    \includegraphics[width=\textwidth]{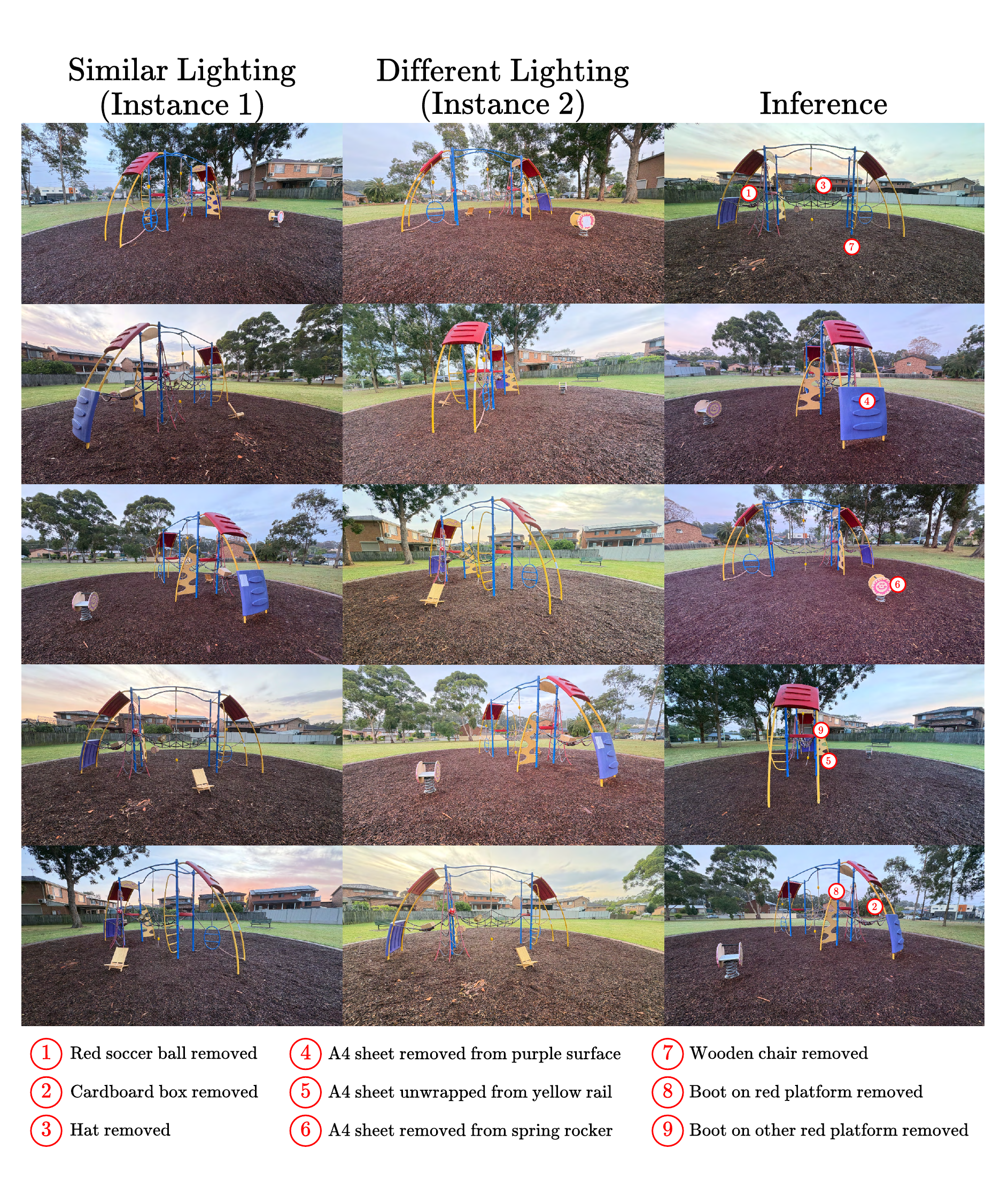}
    \caption{Playground scene visualizations and change descriptions.}
    \label{fig:supp_playground}
\end{figure*}

\begin{figure*}[t]
    \centering
    \includegraphics[width=\textwidth]{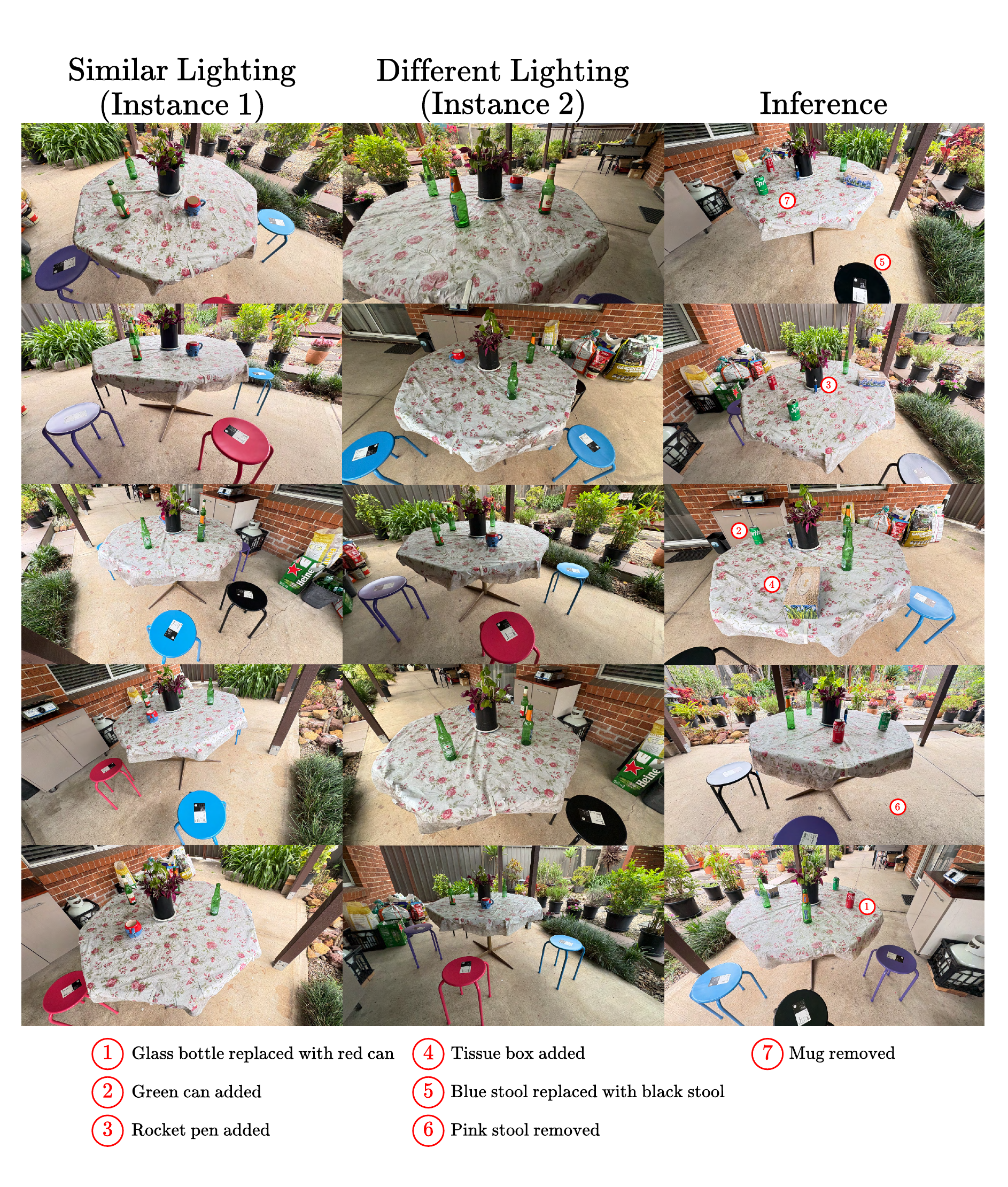}
    \caption{Porch scene visualizations and change descriptions.}
    \label{fig:supp_porch}
\end{figure*}

We also provide additional visualizations and a description of the changes for our PASLCD dataset for each scene at the end of this Supp. Material: Cantina (see Fig.~\ref{fig:supp_cantina}), Lounge (see Fig.~\ref{fig:supp_lounge}), Printing area (see Fig.~\ref{fig:supp_printing}), Lunch Room (see Fig.~\ref{fig:supp_lunch}), Meeting Room (see Fig.~\ref{fig:supp_meeting}), Garden (see Fig.~\ref{fig:supp_garden}), Pots (see Fig.~\ref{fig:supp_pots}), Zen (see Fig.~\ref{fig:supp_zen}), Playground (see Fig.~\ref{fig:supp_playground}) and Porch (see Fig.~\ref{fig:supp_porch}).

\begin{figure}[tb]
    \centering
    % First subfigure: Per-Image Change Area Distribution
    \begin{subfigure}{1.\linewidth}
        \centering
        \begin{tikzpicture}
        \begin{axis}[
            ybar stacked,
            ybar legend,
            ymin=0,
            ymax=160,
            xmin=0,
            xmax=21,
            xtick={5,10,15,20},
            ylabel={Number of Images},
            xlabel={Change Area (\%)},
            title={Per-Image Change Area Distribution},
            legend style={at={(0.73,0.98)}, anchor=north, legend columns=-1},
            bar width=8pt,
            grid=major,  
            major grid style={dashed,gray}
        ]
        
        % Indoor data (stacked on top of outdoor)
        \addplot+[
          ybar
        ] table [x=bin_edge, y=indoor] {change_ratios.dat};
        
        % Outdoor data (stacked on top of indoor)
        \addplot+[
          ybar
        ] table [x=bin_edge, y=outdoor] {change_ratios.dat};

        % Legend for the different data sets
        \legend{Indoor, Outdoor}
        \end{axis}
        \end{tikzpicture}
        \caption{}
    \end{subfigure}\\
    
    % Second subfigure: Change Type Distribution
    \begin{subfigure}{1.\linewidth}
        \centering
        \begin{tikzpicture}
        \begin{axis}[
            ybar stacked,
            symbolic x coords={Add, Remove, Move/Swap},
            xtick=data,
            ymin=0,
            ymax=45,
            enlarge x limits=0.3,
            ylabel={Frequency},
            xlabel={Change Type},
            grid=major,
            major grid style={dashed,gray},
            bar width=50pt,
            legend style={at={(0.72,0.98)}, anchor=north, legend columns=-1},
            title={Change Type Distribution}
        ]
        
        % Load data from the file
        \addplot table[x=Change,y=Structural]{change_type.dat};
        \addplot table[x=Change,y=Surface]{change_type.dat};
        
        \legend{Structural, Surface}
        \end{axis}
        \end{tikzpicture}
        \caption{}
    \end{subfigure}
    
    \caption{PASLCD dataset statistics. (a) Percentage of changed pixels across all images. (b) Distribution of change types, including structural (struct.) and surface (surf.) changes.}
    \label{fig:supp_paslcd_stats}
\end{figure}

\section{Additional Experimental Results}

\begin{figure*}[h]
    \centering
    \includegraphics[width=\textwidth]{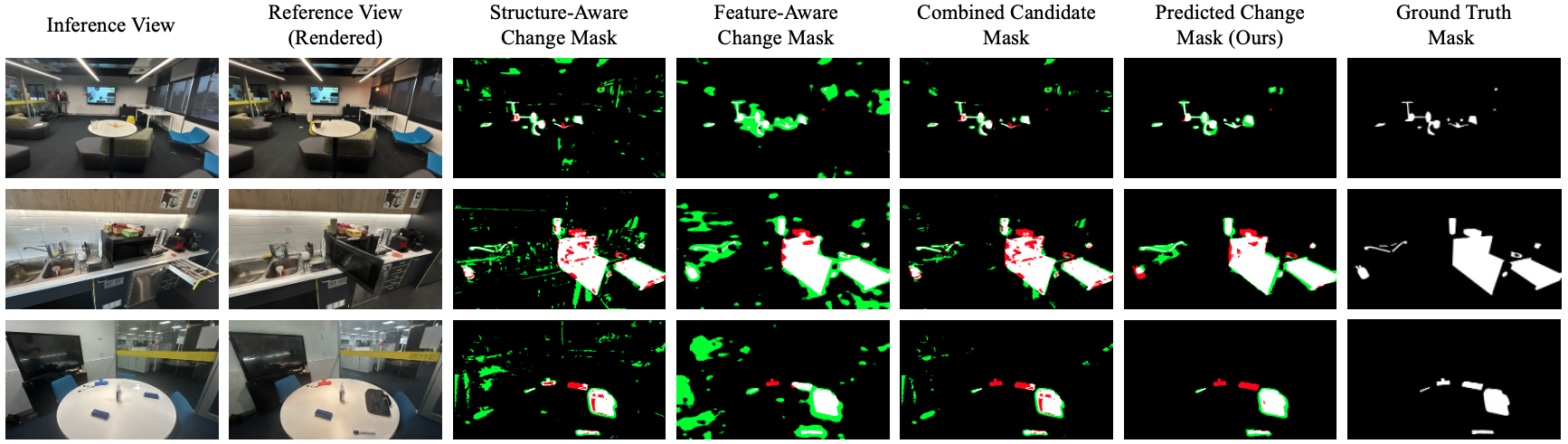}    \caption{Qualitative visualization of change masks across two instances (under similar/different lighting conditions). From left to right: the inference view, the rendered reference view, the structure-aware change mask, the feature-aware change mask, the combined candidate mask, our predicted change mask, and the ground truth mask. The combined candidate mask effectively suppresses the distractor changes which are likely FPs (in green) by merging complementary information in structural and feature-aware masks, while our predicted change mask further refines the detection by suppressing false positives and aligning closely with the ground truth. The last row illustrates false negative failure cases discussed in Sec.~\ref{sec:comp} (in red). Specifically, the color change in the T-shaped structure goes undetected in the feature-aware mask, while the laminated white paper on the white table is missed in the structure-aware mask, resulting in incomplete change detection.}
    \label{fig:supp_mask}
\end{figure*}

\subsection{Instance-level Results for PASLCD}
Tabs.~\ref{tab:supp_main_1_formatted} and~\ref{tab:supp_main_2_formatted} show per-scene quantitative results for our PASLCD dataset under similar lighting conditions and different lighting conditions respectively. We consistently improve the change localization performance over all the baselines under both settings. 

\begin{figure*}
    \centering
    \includegraphics[width=\textwidth]{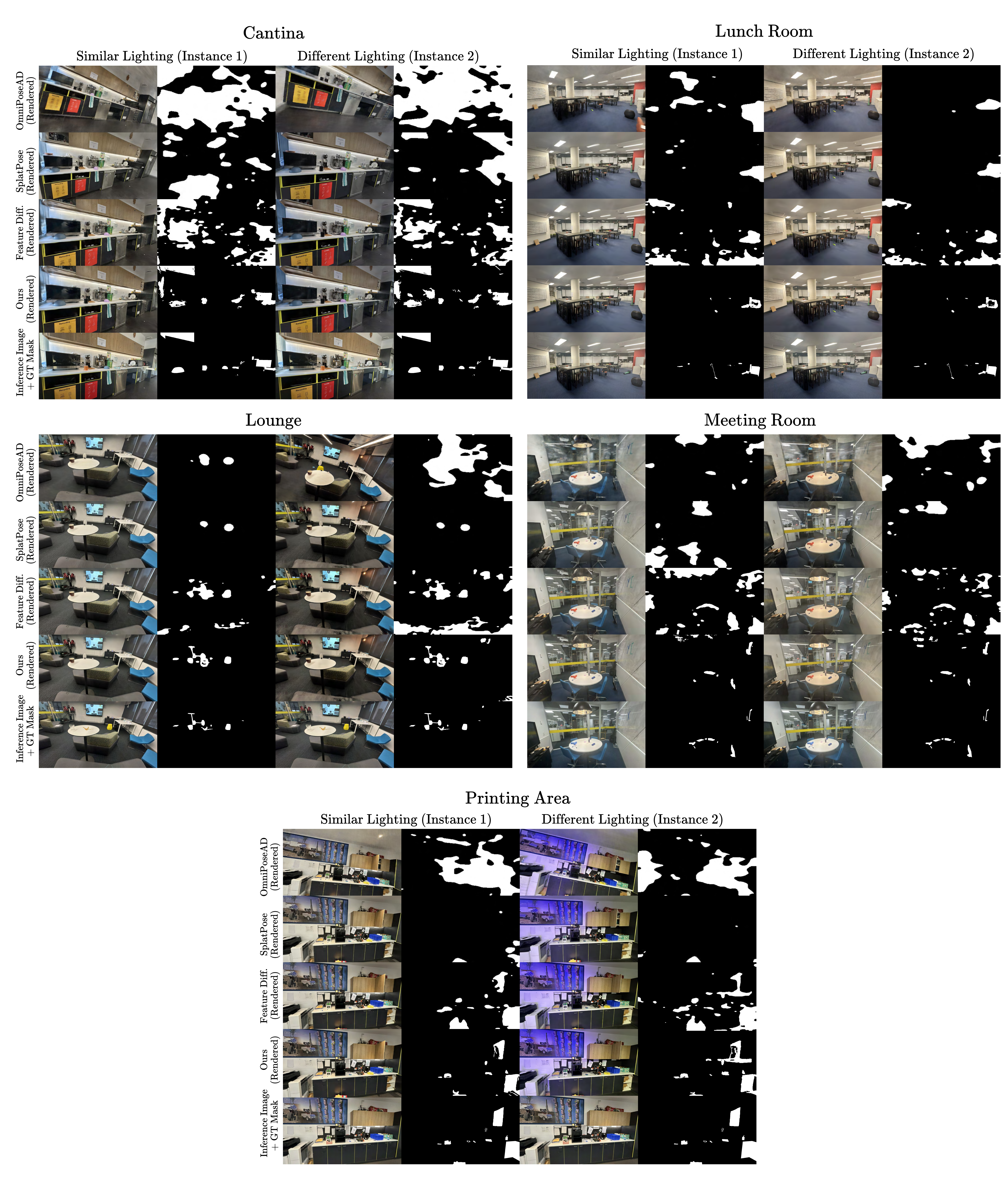}
    \caption{Qualitative results of each method for the indoor scenes of our dataset PASLCD.}
    \label{fig:supp_ours_vis_indoor}
\end{figure*}

\begin{figure*}
    \centering
    \includegraphics[width=\textwidth]{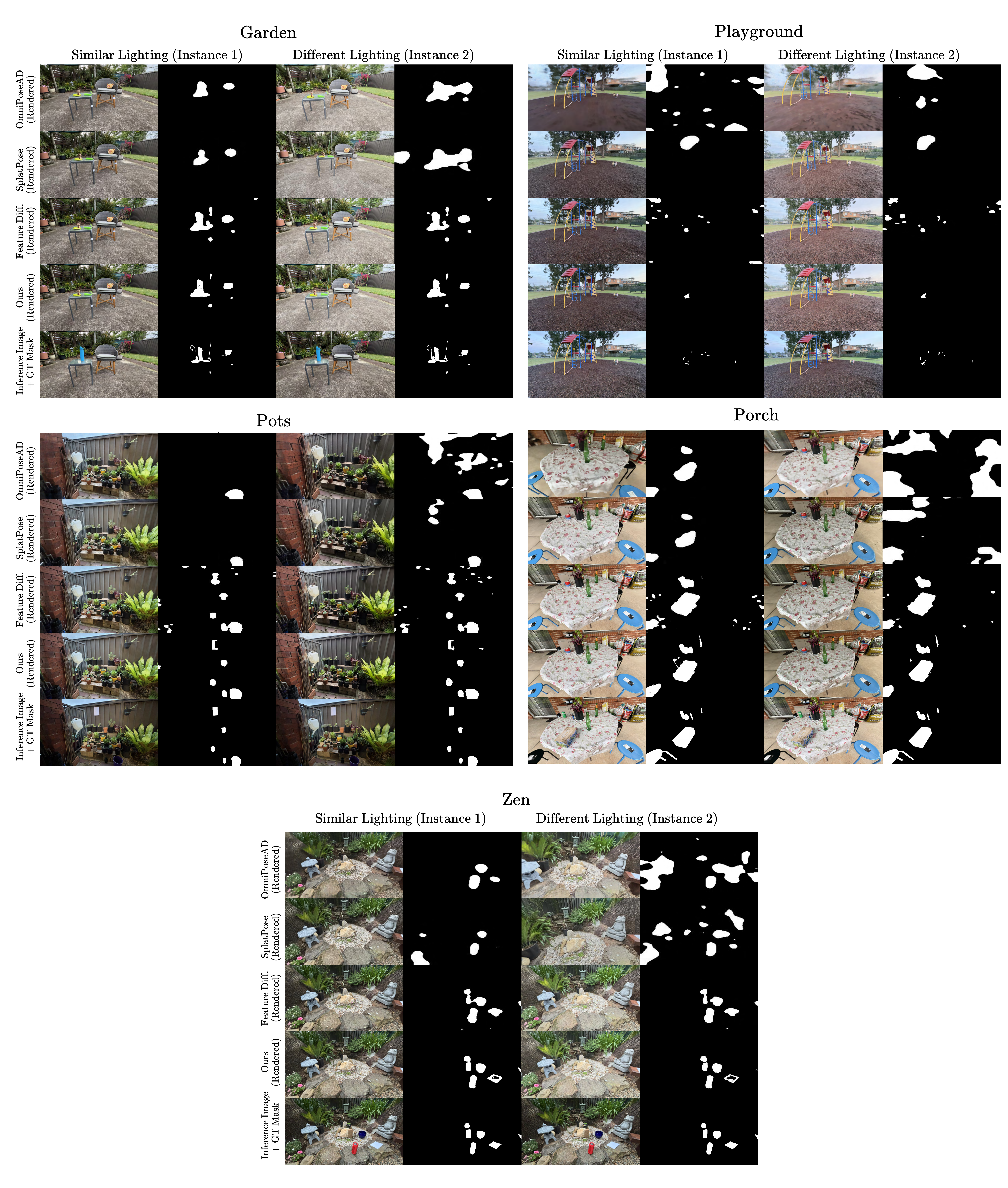}
    \caption{Qualitative results of each method for the outdoor scenes of our dataset PASLCD.}
    \label{fig:supp_ours_vis_outdoor}
\end{figure*}

In Figs.~\ref{fig:supp_ours_vis_indoor} and \ref{fig:supp_ours_vis_outdoor} (placed towards the end of Supp. Material due to size), we show additional qualitative results for all of the methods on PASLCD under the two lighting settings.

\begin{table}[t]
    \centering
    \caption{Relative performance loss (\(\Delta\)) of each method when detecting changes in scenes with different lighting conditions.}
    \label{tab:instance_ablation}
    \scalebox{0.9}{ % Adjust the scaling factor as needed
        \begin{tabular}{lcc} \toprule
            Method & $\Delta$mIoU (\%) $\downarrow$ & $\Delta$F1 (\%) $\downarrow$ \\ \midrule
            CYWS-2D~\cite{sachdeva_change_2023-1} & 16.1 & 10.0 \\
            Feature Diff. & 17.2 & 12.6 \\
            Ours & \textbf{7.2} & \textbf{4.5} \\ \bottomrule
        \end{tabular}
    }
\end{table}

\subsection{Robustness to Distractor Visual Changes:}

In Tab.~\ref{tab:instance_ablation}, we report the relative \emph{loss in performance} of each method (methods having overall mIoU \(\geq 0.2\)) when evaluating under different lighting conditions versus consistent lighting conditions. For both the mIoU and F1 metrics, our multi-view change masks exhibit the least performance drop under different lighting conditions, demonstrating our robustness to distractor visual changes.

\subsection{Complementary Information in Feature-Aware and Structure-Aware Masks}
\label{sec:comp}
In Fig.~\ref{fig:supp_mask}, we illustrate how combining structure-aware and feature-aware masks produces a more effective candidate mask by suppressing likely false positives. The structure-aware and feature-aware masks capture complementary information about false positive change predictions, as shown in the 3rd and 4th columns of Fig.~\ref{fig:supp_mask}. While the feature-aware mask often captures changes as blobs (over-inflating the size of the change) due to the patch-to-pixel interpolation, the structure-aware mask captures more refined change details. However the structure-aware mask suffers from its own false-positive predictions, often due to the edges of fine structures in the scene or due to reflections. Combining both masks together reduces these false change predictions in the candidate mask (see the 5th column in Fig.~\ref{fig:supp_mask}). 

However, as discussed in Sec.~\ref{sec:results}, if one of the masks fails to detect a change, it may result in missing the true change. For instance, in the 3rd row of Fig.~\ref{fig:supp_mask}, the feature-aware mask fails to capture the color change in the T-shaped structure despite the structure-aware mask flagging it, leading to an inability to fully detect the change. This highlights a potential avenue for future research: addressing the limitations of feature masks derived from pre-trained foundation models and effectively leveraging complementary information to produce a more refined change mask.

\begin{table*}[t]
    \centering
    \caption{Quantitative results for our PASLCD dataset, under similar lighting conditions, averaged across \colorbox[HTML]{EAEAEA}{Indoor} and \colorbox[HTML]{F0F8FF}{Outdoor} scenes. The best values per scene are \textbf{bolded}.}
    \label{tab:supp_main_1_formatted}
    \scalebox{0.83}{
        \begin{tabular}{lcccccccccccccc}
            \toprule
            \multirow{2}{*}{Scene} & \multirow{2}{*}{FF/360} & \multicolumn{2}{c}{OmniPoseAD~\cite{zhou2024pad}} & \multicolumn{2}{c}{SplatPose~\cite{kruse2024splatpose}} & \multicolumn{2}{c}{CSCDNet~\cite{sakurada_weakly_2020}} & \multicolumn{2}{c}{CYWS-2D~\cite{sachdeva_change_2023-1}} & \multicolumn{2}{c}{Feature Diff.} & \multicolumn{2}{c}{Ours} \\
            \cmidrule(lr){3-4} \cmidrule(lr){5-6} \cmidrule(lr){7-8} \cmidrule(lr){9-10} \cmidrule(lr){11-12} \cmidrule(lr){13-14}
            & &  mIoU $\uparrow$ & F1 $\uparrow$ & mIoU $\uparrow$ & F1 $\uparrow$ & mIoU $\uparrow$ & F1 $\uparrow$ & mIoU $\uparrow$ & F1 $\uparrow$ & mIoU $\uparrow$ & F1 $\uparrow$ & mIoU $\uparrow$ & F1 $\uparrow$ \\
            \midrule
            \rowcolor[HTML]{EAEAEA} Cantina & FF & 0.146 & 0.239 & 0.210 & 0.333 & 0.088 & 0.151 & 0.296 & 0.434 & 0.351 & 0.506 & \textbf{0.591} & \textbf{0.737} \\
            \rowcolor[HTML]{EAEAEA} Lounge & FF & 0.137 & 0.224 & 0.266 & 0.418 & 0.200 & 0.325 & 0.247 & 0.379 & 0.198 & 0.323 & \textbf{0.498} & \textbf{0.658} \\
            \rowcolor[HTML]{EAEAEA} Printing Area & FF & 0.135 & 0.217 & 0.184 & 0.292 & 0.147 & 0.246 & 0.448 & 0.600 & 0.498 & 0.648 & \textbf{0.637} & \textbf{0.771} \\
            \rowcolor[HTML]{EAEAEA} Lunch Room & 360 & 0.144 & 0.224 & 0.146 & 0.234 & 0.037 & 0.065 & 0.108 & 0.183 & 0.103 & 0.176 & \textbf{0.395} & \textbf{0.551} \\
            \rowcolor[HTML]{EAEAEA} Meeting Room & 360 & 0.095 & 0.168 & 0.156 & 0.247 & 0.208 & 0.325 & 0.145 & 0.246 & 0.128 & 0.222 & \textbf{0.371} & \textbf{0.531} \\
            \midrule
            \rowcolor[HTML]{F0F8FF} Garden & FF & 0.297 & 0.440 & 0.228 & 0.357 & 0.245 & 0.389 & 0.347 & 0.510 & 0.265 & 0.410 & \textbf{0.415} & \textbf{0.578} \\
            \rowcolor[HTML]{F0F8FF} Pots & FF & 0.207 & 0.317 & 0.119 & 0.314 & 0.021 & 0.039 & 0.400 & 0.554 & 0.448 & 0.606 & \textbf{0.569} & \textbf{0.717} \\
            \rowcolor[HTML]{F0F8FF} Zen & FF & 0.232 & 0.352 & 0.192 & 0.304 & 0.009 & 0.016 & 0.455 & 0.554 & 0.454 & 0.586 & \textbf{0.533} & \textbf{0.659} \\
            \rowcolor[HTML]{F0F8FF} Playground & 360 & 0.074 & 0.121 & 0.096 & 0.155 & 0.131 & 0.213 & 0.054 & 0.100 & 0.041 & 0.078 & \textbf{0.244} & \textbf{0.371} \\
            \rowcolor[HTML]{F0F8FF} Porch & 360 & 0.292 & 0.417 & 0.312 & 0.462 & 0.172 & 0.286 & 0.455 & 0.619 & 0.403 & 0.565 & \textbf{0.530} & \textbf{0.688} \\
            \midrule
            Average & -- & 0.176 & 0.272 & 0.191 & 0.312 & 0.126 & 0.206 & 0.295 & 0.418 & 0.289 & 0.412 & \textbf{0.478} & \textbf{0.626} \\
            \bottomrule
        \end{tabular}
    }
\end{table*}

\begin{table*}[t]
    \centering
    \caption{Quantitative results for our PASLCD dataset, under different lighting conditions, averaged across \colorbox[HTML]{EAEAEA}{Indoor} and \colorbox[HTML]{F0F8FF}{Outdoor} scenes. The best values per scene are \textbf{bolded}.}
    \label{tab:supp_main_2_formatted}
    \scalebox{0.83}{
        \begin{tabular}{lcccccccccccccc}
            \toprule
            \multirow{2}{*}{Scene} & \multirow{2}{*}{FF/360} & \multicolumn{2}{c}{OmniPoseAD~\cite{zhou2024pad}} & \multicolumn{2}{c}{SplatPose~\cite{kruse2024splatpose}} & \multicolumn{2}{c}{CSCDNet~\cite{sakurada_weakly_2020}} & \multicolumn{2}{c}{CYWS-2D~\cite{sachdeva_change_2023-1}} & \multicolumn{2}{c}{Feature Diff.} & \multicolumn{2}{c}{Ours} \\
            \cmidrule(lr){3-4} \cmidrule(lr){5-6} \cmidrule(lr){7-8} \cmidrule(lr){9-10} \cmidrule(lr){11-12} \cmidrule(lr){13-14}
            & &  mIoU $\uparrow$ & F1 $\uparrow$ & mIoU $\uparrow$ & F1 $\uparrow$ & mIoU $\uparrow$ & F1 $\uparrow$ & mIoU $\uparrow$ & F1 $\uparrow$ & mIoU $\uparrow$ & F1 $\uparrow$ & mIoU $\uparrow$ & F1 $\uparrow$ \\
            \midrule
            \rowcolor[HTML]{EAEAEA} Cantina & FF & 0.130 & 0.222 & 0.166 & 0.274 & 0.069 & 0.124 & 0.259 & 0.383 & 0.151 & 0.258 & \textbf{0.569} & \textbf{0.720} \\
            \rowcolor[HTML]{EAEAEA} Lounge & FF & 0.161 & 0.258 & 0.257 & 0.402 & 0.189 & 0.311 & 0.196 & 0.317 & 0.156 & 0.269 & \textbf{0.428} & \textbf{0.593} \\
            \rowcolor[HTML]{EAEAEA} Printing Area & FF & 0.179 & 0.267 & 0.181 & 0.283 & 0.147 & 0.245 & 0.206 & 0.314 & 0.366 & 0.520 & \textbf{0.539} & \textbf{0.697} \\
            \rowcolor[HTML]{EAEAEA} Lunch Room & 360 & 0.177 & 0.269 & 0.119 & 0.196 & 0.033 & 0.059 & 0.137 & 0.226 & 0.099 & 0.172 & \textbf{0.382} & \textbf{0.540} \\
            \rowcolor[HTML]{EAEAEA} Meeting Room & 360 & 0.118 & 0.196 & 0.104 & 0.175 & 0.218 & 0.335 & 0.130 & 0.220 & 0.115 & 0.200 & \textbf{0.328} & \textbf{0.483} \\
            \midrule
            \rowcolor[HTML]{F0F8FF} Garden & FF & 0.249 & 0.382 & 0.141 & 0.243 & 0.236 & 0.377 & 0.346 & 0.508 & 0.318 & 0.479 & \textbf{0.456} & \textbf{0.623} \\
            \rowcolor[HTML]{F0F8FF} Pots & FF & 0.079 & 0.142 & 0.161 & 0.267 & 0.023 & 0.042 & 0.301 & 0.508 & 0.346 & 0.525 & \textbf{0.510} & \textbf{0.669} \\
            \rowcolor[HTML]{F0F8FF} Zen & FF & 0.255 & 0.375 & 0.179 & 0.288 & 0.010 & 0.019 & 0.445 & 0.596 & 0.434 & 0.568 & \textbf{0.466} & \textbf{0.606} \\
            \rowcolor[HTML]{F0F8FF} Playground & 360 & 0.078 & 0.129 & 0.066 & 0.111 & 0.137 & 0.224 & 0.065 & 0.115 & 0.052 & 0.099 & \textbf{0.254} & \textbf{0.384} \\
            \rowcolor[HTML]{F0F8FF} Porch & 360 & 0.255 & 0.375 & 0.166 & 0.263 & 0.180 & 0.297 & 0.423 & 0.595 & 0.354 & 0.511 & \textbf{0.505} & \textbf{0.664} \\
            \midrule
            Average & -- & 0.160 & 0.252 & 0.154 & 0.250 & 0.124 & 0.203 & 0.251 & 0.378 & 0.239 & 0.360 & \textbf{0.444} & \textbf{0.598} \\
            \bottomrule
        \end{tabular}
    }
\end{table*}

\end{document}